%% file: llameabo_arxiv_clean.tex
\documentclass{article}

\usepackage[preprint]{neurips_2025}

\usepackage[utf8]{inputenc} 
\usepackage[T1]{fontenc}    
\usepackage{hyperref}       
\usepackage{url}            
\usepackage{booktabs}       
\usepackage{amsfonts}       
\usepackage{nicefrac}       
\usepackage{microtype}      
\usepackage{xcolor}         
\usepackage{amsmath}
\usepackage{graphicx} 
\usepackage{algorithm}
\usepackage{algorithmicx}
\usepackage{algpseudocode}
\usepackage{framed}
\usepackage[most]{tcolorbox}
\usepackage{placeins}
\usepackage{listings}
\usepackage{caption}
\usepackage{subcaption}

\usepackage{fontawesome5}

\lstdefinestyle{mintedstyle}{
	language=Python,
	basicstyle=\ttfamily\scriptsize,
	keywordstyle=\color{blue}\bfseries,
	commentstyle=\color{gray}\itshape,
	stringstyle=\color{teal},
	numberstyle=\tiny\color{gray},
	numbers=left,
	numbersep=8pt,
	backgroundcolor=\color{gray!5},
	frame=single,
	rulecolor=\color{gray},
	breaklines=true,
	showstringspaces=false,
	tabsize=2,
	captionpos=b
}

\definecolor{peachpink}{rgb}{0.75, 0.95, 0.75}

\newtcolorbox{notebox}{
  colback=peachpink,
  colframe=peachpink,
  boxrule=0pt,
  arc=4pt,
  left=4pt,
  right=4pt,
  top=4pt,
  bottom=4pt,
  enhanced jigsaw,
  before skip=6pt,
  after skip=6pt,
  fontupper=\itshape,
  overlay={
    \node[anchor=north west, inner sep=5pt] at (frame.north west) {\faExclamationCircle};
  }
}

\tcbset {
  base/.style={
    arc=0mm, 
    bottomtitle=0.5mm,
    boxrule=0mm,
    colbacktitle=black!10!white, 
    coltitle=black, 
    fonttitle=\bfseries, 
    left=2.5mm,
    leftrule=1mm,
    right=3.5mm,
    title={#1},
    toptitle=0.75mm, 
    breakable
  }
}

\definecolor{brandblue}{rgb}{0.34, 0.7, 1}
\newtcolorbox{mainbox}[1]{
  colframe=brandblue, 
  base={#1}
}

\newtcolorbox{subbox}[1]{
  colframe=black!30!white,
  base={#1}
}

\usepackage{tikz}
\usetikzlibrary{fit,calc}
\newcommand*{\tikzmk}[1]{\tikz[remember picture,overlay,] \node (#1) {};\ignorespaces}
\newcommand{\boxit}[1]{\tikz[remember picture,overlay]{\node[yshift=0pt,fill=#1,opacity=.15,fit={($(A)-(0.7,0.2)$)($(B)+(.95\linewidth,0.8\baselineskip)$)}] {};}\ignorespaces}
\colorlet{myyellow}{yellow!40}
\colorlet{mypink}{red!40}
\colorlet{mycyan}{blue!40}

\title{LLaMEA-BO: A Large Language Model Evolutionary Algorithm for Automatically Generating Bayesian Optimization Algorithms}

\author{%
  Wenhu Li, \ Niki van Stein, \ Thomas B\"{a}ck, \ and \ Elena Raponi \\
  Leiden Institute of Advanced Computer Science\\
  Leiden University\\
  Leiden, The Netherlands \\
  \texttt{w.li.12@umail.leidenuniv.nl} 
}

\begin{document}

\maketitle

\begin{abstract}

Bayesian optimization (BO) is a powerful class of algorithms for optimizing expensive black-box functions, but designing effective BO algorithms remains a manual, expertise-driven task.
Recent advancements in Large Language Models (LLMs) have opened new avenues for automating scientific discovery, including the automatic design of optimization algorithms.
While prior work has used LLMs within optimization loops or to generate non-BO algorithms, we tackle a new challenge: Using LLMs to automatically generate full BO algorithm code. 
Our framework uses an evolution strategy to guide an LLM in generating Python code that preserves the key components of BO algorithms: An initial design, a surrogate model, and an acquisition function. The LLM is prompted to produce multiple candidate algorithms, which are evaluated on the established Black-Box Optimization Benchmarking (BBOB) test suite from the COmparing Continuous Optimizers (COCO) platform. 
Based on their performance, top candidates are selected, combined, and mutated via controlled prompt variations, enabling iterative refinement.
Despite no additional fine-tuning, the LLM-generated algorithms outperform state-of-the-art BO baselines in 19 (out of 24) BBOB functions in dimension 5 and generalize well to higher dimensions, and different tasks (from the Bayesmark framework). This work demonstrates that LLMs can serve as algorithmic co-designers, offering a new paradigm for automating BO development and accelerating the discovery of novel algorithmic combinations. The source code is provided at \url{https://github.com/Ewendawi/LLaMEA-BO}.
    
\end{abstract}

\section{Introduction} 
Bayesian optimization (BO) \citep{frazierTutorialBayesianOptimization2018,garnettBayesianOptimizationBook} is a widely used optimization framework for optimizing expensive black-box functions, i.e., functions that lack an analytical form and for which derivative information is unavailable. 
To do this, BO iteratively selects high-utility points using a surrogate model---commonly a Gaussian process (GP) under continuous settings---and an acquisition function to efficiently explore and exploit the search space.
Over the past decades, BO has become one of the most actively researched and widely applied frameworks for tackling optimization problems with expensive evaluations, particularly for applications in machine learning \citep{snoekPracticalBayesianOptimization2012a,turnerBayesianOptimizationSuperior2021c} and algorithm selection and configuration \citep{lindauerSMAC3VersatileBayesian2022b}. Here, tuning hyperparameters of models like neural networks, optimization algorithms, or ensemble methods often involves expensive training and validation procedures, which can be prohibitively slow or computationally expensive. As a result, BO has emerged as the de facto standard for hyperparameter optimization \citep{feurerHyperparameterOptimization2019a}, and has also found applications in areas such as experimental design, automated machine learning, simulation-based optimization, and control. 

The success of BO in these domains has brought many methodological advances, including the development of new acquisition functions \citep{frazierKnowledgeGradientPolicyCorrelated2009,wangNewAcquisitionFunction2017a,deathGreedGoodExploration2021c,hvarfnerPBOAUGMENTINGACQUISITION2022,hvarfnerJointEntropySearch2023a}, surrogate models beyond GPs \citep{breimanRandomForests2001,geurtsExtremelyRandomizedTrees2006,jiangEnsemblesSurrogateModels2020,hoofHyperboostHyperparameterOptimization2021}, handling of constraints \citep{gardnerBayesianOptimizationInequality2014a,hernandez-lobatoGeneralFrameworkConstrained2016a,ungreddaBayesianOptimisationConstrained2021a,erikssonScalableConstrainedBayesian2021c}, scalability to high-dimensional problems \citep{erikssonScalableGlobalOptimization2020,erikssonHighDimensionalBayesianOptimization2021,ziomekAreRandomDecompositions2023b,santoniComparisonHighDimensionalBayesian2023,papenmeierIncreasingScopeYou2023}, and integration with parallel and asynchronous evaluation \citep{gonzalezBatchBayesianOptimization2015,alviAsynchronousBatchBayesian2019,marminDifferentiatingMultipointExpected2019}. 
Yet, designing effective BO algorithms still remains an intensive process relying on domain-expert knowledge to select and tune main components such as the initial design scheme, the choice of a surrogate model, or an acquisition strategy to determine what parameter configurations are convenient to evaluate. 

This highly modular framework characterizing BO makes it a perfect candidate for automatic algorithm design and the recent advances in large language models (LLMs) offer a new lens through which that can be done. 
LLMs have shown remarkable abilities in zero-shot code generation, few-shot reasoning, and even synthesizing novel scientific pipelines. Building on this trend, in this work we explore whether LLMs can autonomously generate complete and functioning BO algorithms in code, moving beyond the usage of LLMs to assist in single components of the modular framework \citep{ramosBayesianOptimizationCatalysts2023,yinADOLLMAnalogDesign2024,yaoEvolveCostawareAcquisition2024}. 

\textbf{Our contribution.} Our approach builds on the Large-Language-Model Evolutionary Algorithm (LLaMEA) framework introduced by \citet{stein2025llamea}, which treats LLMs as algorithmic generators in a closed-loop generate–evaluate–improve cycle. While the original LLaMEA uses a $(1+1)$ evolution strategy (ES) \citep{schwefelEvolutionOptimumSeeking1993} to optimize LLM-generated code for metaheuristics, we present a new variant that adapts this idea to the domain of BO. Our primary contributions are:
\begin{itemize}
    \item We propose \textbf{LLaMEA-BO}, which utilizes multiple ES variants and introduces BO-specific prompt structures along with dedicated crossover and mutation operators to evolve increasingly effective algorithms.
    \item We \textbf{validate} the quality of the best BO algorithms generated by LLaMEA-BO through \textbf{extensive benchmarking} against state-of-the-art BO baselines on the BBOB test suite from the COCO platform, testing their generalization capabilities across multiple dimensions, functions, and instances, and further assess robustness on unseen tasks using the Bayesmark framework. 
    \item We provide the code of the best generated algorithm and analyze its performance through \textbf{extensive ablation studies} on its hyperparameters.
\end{itemize}

\begin{notebox}
    \quad \  LLaMEA-BO is \textbf{the first} LLM-driven framework to automatically evolve \textbf{complete} BO \\ algorithms which are \textbf{outperforming} current state-of-the-art BO baselines.
\end{notebox}

\section{Related Work}
\textbf{LLMs for Code Generation.} 
LLMs are increasingly employed as \emph{generative engines} that write and iteratively refine source code, enabling the automated discovery of optimization heuristics or the fine-tuning of existing algorithms \citep{Pluhacek2024,liu2024systematicsurveylargelanguage}.  A prototypical workflow for the discovery of new (meta)heuristics couples an LLM with a search strategy that evaluates each candidate on a downstream task and feeds performance signals back to the model.  Notable instances include \emph{FunSearch}, which explores program space for combinatorial problems such as the cap‑set and bin‑packing tasks \citep{FunSearch2024}; AEL, which evolves code snippets via LLM‑mediated crossover and mutation \citep{liu2023algorithm}; and \emph{Evolution of Heuristics} (EoH), where every program is a population member that is mutated by the LLM \citep{fei2024eoh}.  While these approaches yield promising results on small TSP instances, scalability to harder or continuous problems remains limited.

\cite{zhang2024understanding} systematically benchmarked LLM‑centred evolutionary program search (EPS) methods and confirmed that a dedicated search component is essential as stand‑alone prompting is rarely competitive.  Their study also highlighted pronounced variance across LLM back‑ends and problem domains.

The LLaMEA framework \citep{stein2025llamea,stein2025llameahpo} extends this line of work to continuous black‑box optimisation: an LLM generates complete Python classes, including stateful components, that are evaluated with \textsc{IOHexperimenter} \citep{IOHexperimenter}. Selection, mutation prompts, and automatic error handling close the loop, allowing the discovery of metaheuristics that outperform state-of-the-art optimization algorithms such as CMA‑ES and Differential Evolution on the BBOB testbed \citep{hansen2009real}.

\textbf{Bayesian Optimization} 
Bayesian Optimization (BO) \citep{garnettBayesianOptimizationBook} is a model-based approach for sample-efficient optimization of expensive black-box functions, where each evaluation is costly. In continuous domains, a Gaussian Process (GP) is commonly used to model the unknown objective, encoding prior assumptions about smoothness and variability.
The procedure typically starts with an initial set of evaluated points---commonly referred to as a Design of Experiments \citep{forresterEngineeringDesignSurrogate2008}---which are selected through random sampling or space-filling methods.
Once this initial data is gathered, the GP is conditioned on the observations to produce a posterior distribution. This posterior not only estimates the objective function but also quantifies uncertainty across the search space.
An acquisition function (AF) then guides the selection of the next evaluation point by balancing two competing goals: exploring uncertain regions and exploiting areas likely to yield high values.
This cycle of modeling, selection, and evaluation is repeated iteratively until the allocated budget is used up or a satisfactory solution is found.

\textbf{LLMs in Bayesian Optimization} 
Recent works have explored the use of LLMs in BO.
\cite{ramosBayesianOptimizationCatalysts2023} proposed a zero-shot BO method using LLMs that predict catalyst properties directly from natural language synthesis descriptions, enabling molecule design without the need for model training or feature engineering. The authors use LLMs as in-context surrogate regressor that predicts catalyst properties from natural language synthesis procedures, also deriving mean and uncertainty to allow for the definition of an acquisition function.
\cite{yinADOLLMAnalogDesign2024} proposed the ADO-LLM framework, which integrates LLMs with BO to address the complex multi-objective optimization challenges inherent in analog circuit sizing. They use LLMs for both the zero-shot initialization phase to suggest viable design points and the generation of new candidate solutions in the acquisition phase via in-context learning.
\cite{yaoEvolveCostawareAcquisition2024} introduced a novel framework that integrates LLMs with evolutionary computation to automatically design cost-aware acquisition functions through iterative crossover and mutation in the algorithmic space (an acquisition function is defined by its algorithm description and code block implementation). 
All the preceding examples use LLMs to improve specific components of the BO pipeline: improving initial sampling, serving as surrogate models, informing the acquisition function, or a combination of these. It is with the most recent work by \cite{liuLargeLanguageModels2024a} that LLMs are used in all phases of a the modular BO framework (framed in natural language), proving to improve the effectiveness of the algorithm compared to established algorithms, especially in the early stages of the search when observations are sparse. 

Despite recent advances, LLMs have not yet been used to generate complete Python implementations of Bayesian optimization (BO) algorithms, as LLaMEA \citep{stein2025llamea} does for evolutionary algorithms. We address this gap by introducing LLaMEA-BO, an extension of LLaMEA that generates full BO algorithms by prompting an LLM to produce implementations that include all the essential components of a BO pipeline: initialization, surrogate modeling, acquisition function optimization, and candidate evaluation.

\section{LLaMEA-BO}
\label{sec:methodology}

\emph{The Large‑Language‑Model Evolutionary Algorithm (LLaMEA)}~\citep{stein2025llamea} is an
automated \emph{generate–evaluate–improve} loop in which an LLM proposes the
source code of a candidate optimization algorithm, that code is compiled and
benchmarked, and the performance signal (plus any error information) is fed back into
the next prompt.  The original LLaMEA runs a minimalist
\mbox{$(1\!+\!1)$} evolution strategy---one parent and one child per
iteration---which keeps LLM‑query cost low and progress interpretable.

The Bayesian‑optimisation flavour introduced here, \textbf{LLaMEA‑BO},
preserves the same evolution‑through‑prompting philosophy but extends it to a
\textbf{population‑based} algorithm with dedicated crossover operators tailored
to BO.  Algorithm~\ref{alg:LLaMEABO} gives an overview of the methodology; the remainder of this section details the methodological differences with the original approach.

\input{llameabo_pseudocode}

\subsection{Task Prompt and BO‑specific Template}
\label{sec:prompt}

Each call to the LLM starts with a \textit{task prompt}~$S$ that \textbf{(1)} declares the model’s \textbf{role}: ``design a novel and efficient Bayesian optimization algorithm''; \textbf{(2)} embeds a \textbf{working template} that already contains the canonical BO elements such as; initial design, surrogate, acquisition function, and outer optimization loop; and \textbf{(3)} demands a \texttt{Python} \textbf{class} whose \verb|__call__(self, f, budget)| method returns the incumbent~$x^\star$.

The complete template and all prompts are available in our open source repository \cite{LLaMEABO_Anonymous}, raw logs and experiment data are available in our Zenodo repository \cite{anonymous_2025_15384611}. 

\textbf{Difference to earlier variants.} 
Vanilla LLaMEA provided only a  
random‑search skeleton with minimal code instructions. LLaMEA‑BO instead ships a \emph{fully fledged} but \emph{generic} BO loop,
steering the LLM toward innovations in kernels, trust‑region logic, or acquisition scheduling instead of re‑implementing boiler‑plate code; this reduces syntax errors and accelerates convergence. Early experiments with vanilla LLaMEA to produce working BO algorithms (without the BO specific code template) did not perform at all.
Another notable difference is that the initial population is generated with diversity in mind by asking the LLM to generate something different from what has been generated so far using the below prompting template.

\begin{mainbox}{Diversity Initialization Prompt}
You are a highly skilled computer scientist in the field of natural computing. Your task is to design novel metaheuristic algorithms to solve black box optimization problems.\\
\textit{<task prompt>}\\
$n$ algorithms have been designed. 
The new algorithm should be as \textbf{diverse} as possible from the previous ones on every aspect.\\
If errors from the previous algorithms are provided, analyze them. The new algorithm should be designed to avoid these errors.\\
\textit{<list of $n$ algorithms>} \\
\textit{<code format example (Appendix \ref{sec:code-template})>}\\
\textit{<output format specification>}
\end{mainbox}

\subsection{Evaluation}
\label{sec:evaluation}

\textbf{Evaluation during the evolutionary loop. }
Following the protocol of \textsc{LLaMEA}~\citep{stein2025llamea},
we measure performance of all generated algorithms with the \emph{anytime} \textbf{Area Over the Convergence Curve} ($AOCC$)
on a subset of $10$ of the $24$ noiseless functions of COCO's BBOB suite~\cite{hansen2021coco}.
The only substantive change is the much \emph{tighter evaluation budget} of $B= 20\!\times\!d$ function evaluations
($d = 5$ in our study for discovering algorithms, hence $B = 100$),
which is common in BO benchmarks and stresses rapid convergence and exploitation of the surrogate model.
We choose a subset of $10$ functions---function ids $2, 4, 6, 8, 12, 14, 15, 18, 21$ and $23$ (two functions per high-level function class) to be precise---as this lowers the computational time required while it maintains a large variety in function landscapes. In addition, we exclude some BBOB functions as they have known biases in the locations of optima (around the center) which can bias the generated algorithms \citep{long2023bbob}.
Formal definitions of the AOCC performance metric and details about how we aggregate across settings are available in Appendix \ref{appendix:metrics}.

\textbf{Evaluation of final algorithms. }
We benchmark the top generated BO algorithms on the full set of $24$ test functions of BBOB using $3$ different test instances per function and 5 random seeds. In this evaluation step we also test for generalization beyond the initial $5d$ setting to higher dimensions, up to $40d$.
Under these settings, we measure algorithm performance in terms of \textit{loss}, defined as 
$\min_{h \in \mathcal{H}_t} | f(h) - f^*_{\min} |$,
where $\mathcal{H}_t$ denotes the points chosen up to trial $t$, and $f^*_{\min}$ represents the global minimum, available for the BBOB functions.
In addition, we benchmark the BO algorithms on $25$ tasks extracted from \emph{Bayesmark} \citep{UberBayesmark2025}, a benchmark framework to designed to compare BO methods on real machine learning tasks. 
On Bayesmark, a task is a dataset-ML model pair and we measure performance based on the task metric, accuracy for classification tasks, and MSE for regression tasks. Also, in line with the empirical analysis performed in \citep{liuLargeLanguageModels2024a}, we assess the performance of the generated BO algorithms on $3$ additional synthetic datasets generated from complex multimodal functions: Rosenbrock, Grienwank and KTablet \citep{watanabeTreeStructuredParzenEstimator2023a}. These tasks where selected to assess the generalization capabilities of the algorithms generated by LLaMEA-BO to test beds that are not used for in-loop evaluation during automatic algorithm design. On Bayesmark, we measure algorithm performance in terms of \textit{normalized regret} \citep{liuLargeLanguageModels2024a}, defined as $\min_{h \in \mathcal{H}_t} \left( f(h) - s^*_{\min} \right) / \left( s^*_{\max} - s^*_{\min} \right)$,
where $s^*_{\min}, s^*_{\max}$ represent the best and worst scores, measured by accuracy for classification tasks and mean squared error (MSE) for regression tasks.

\subsection{Variation Operators and Population‑based Selection}
\label{sec:variation}

We keep a \emph{population}~$P_t$ of size~$\mu$ and generate
$\lambda$ offspring per generation.
The crossover, mutation and selection procedure is highlighted in Algorithm \ref{alg:LLaMEABO} as it is significantly different from vanilla LLaMEA.

\textbf{Mutation. }
Mutation follows vanilla LLaMEA: a single‑parent prompt appends the parent's source code, its score, and an instruction to improve the selected algorithm.

\begin{subbox}{Mutation Prompt} 
The selected solution to update is:
\textit{<parent $a$>}\\
Refine the strategy of the selected solution to improve it.
\end{subbox}

\textbf{Crossover. }
Because BO implementations are typically concise ($< 250$ lines of code) and highly
\emph{composable}, meaningful crossover becomes effective once we work with a population.
At every generation:

\begin{enumerate}
  \item Sort~$P_t$ by AOCC and keep the top $\lceil \mu/2 \rceil$ as the \emph{mating pool}.
  \item Walk through the cyclic parent list
        \(
          (a_1,a_2),\,(a_1,a_3),\dots,(a_{k-1},a_k)
        \)
        in the mating pool, first paring the best with all other solutions, then the second best and so on, 
        and for each pair issue a \textbf{two‑parent prompt}:

        \begin{mainbox}{Crossover Prompt} 
          The selected solutions to update are: 
          \textit{<parent $a_i$>}
          \textit{<parent $a_j$>}\\
          Combine the selected solutions to create a new solution. Then refine the strategy of the new solution to improve it. If the errors from the previous algorithms are provided, analyze them. The new algorithm should be designed to avoid these errors.
        \end{mainbox}

  \item Based on the crossover rate $p_{cr}$, the LLM returns $\lambda_c$ crossover children and 
  $\lambda_m=\lambda-\lambda_c$ offspring are produced by mutation.
\end{enumerate}

\textbf{Selection. }
After evaluation we apply either (i) \emph{comma‑selection} ($\lambda$, $\mu$): choose the best $\mu$
        offspring from $\lambda$ offspring, discarding all $\mu$ parents (explorative), 
        or
(ii) \emph{plus‑selection} ($\mu\!+\!\lambda$): pick the best $\mu$ from the $\mu$ parents \emph{and} $\lambda$ offspring (elitist).
The mode is controlled by an \texttt{isElitism} flag
(Algorithm~\ref{alg:LLaMEABO}).
Ties are broken in favour of the algorithm with fewer lines of code, thus promoting parsimonious designs.


\section{Experimental Setup}
We conduct experiments to validate the proposed LLaMEA-BO algorithm, benchmark the top-performing generated BO variants, and assess the impact of key hyperparameters through extensive ablation studies (see Appendix~\ref{appendix:ablation} for the ablations). Our core LLM is \textit{gemini-2.0-flash}, selected for its strong performance in prior tests and free API availability in many countries, which supports reproducibility. As LLM parameters, we use a \textit{temperature} of $0.5$ and \textit{Top-k} sampling of $60$; Appendix~\ref{appendix:LLMconfig} provides ablations supporting these choices.

\subsection{Experimental Setup for the LLaMEA-BO Search}
\label{sec:exp-setup-llamea-bo}

We first evaluate the core LLaMEA-BO framework itself.  
Every experiment starts from a \textit{parent} population that the LLM extends generation‑by‑generation by proposing Python source code candidate slutions for new BO algorithms; each proposed solution is executed and scored
in the loop.  The search is configured as a $(\mu\!+\!/\!,\lambda)$
evolution strategy (ES) with three different settings (that are common in ES literature): Elitist $(1\!+\!1)$–ES, elitist $(4\!+\!16)$–ES, and non‑elitist $(8,16)$–ES.

\textbf{Search budget and repetitions.} 
One \emph{search budget unit} equals the
generation and evaluation of one candidate algorithm.  
Each independent run is limited to $100$ such units
and stops when either the budget or a runtime limit of
$30$\,min is reached.  
For every ES setting we perform $5$ independent runs.

\textbf{Benchmark and fitness.} 
Every candidate algorithm is evaluated according to the procedure explained in Section \ref{sec:evaluation}, 
using \textsc{IOH}\,experimenter \citep{IOHexperimenter} on the $10$
BBOB functions \#\{2,4,6,8,12,14,15,18,21,23\} in dimension $d=5$.
Candidate evaluations that raise an exception receive an evaluation score (AOCC) of~$0$.

\subsection{Experimental Setup for Validation}
\label{sec:exp_set_validation}
To validate the generalization of the generated BO algorithms we benchmark the top $3$ generated BO algorithms on two large benchmark suites: BBOB and \emph{BayesMark}, to which $3$ additional synthetic datasets generated from multimodal functions are added, in line with the empirical evaluation in \citep{liuLargeLanguageModels2024a}. This gives a total of $24$ analytical test functions from BBOB (for each function we run $5$ independent repetitions on $3$ different instances $\{4,5,6\}$) and $40$ machine learning tasks from the extended Bayesmark suite.
For BBOB, we consider dimensions $d \in \{5,10,20,40\}$ and compare performance based on convergence curves over a total evaluation budget of $10d+50$. 
For the extended Bayesmark, we consider both regression and classification tasks ranging from $d=2$ to $d=6$, on a total evaluation budget equal to $30$, constant across tasks. 
The state-of-the-art baselines we compare our generated algorithms to are the Covariance Matrix Adaptation Evolution Strategy (CMA-ES) \citep{hansenCMAEvolutionStrategy2023a}, and the well-established BO approaches Heteroscedastic and Evolutionary Bayesian Optimization (HEBO) \citep{cowen-riversHEBOPushingLimits2022}, Trust-Region Bayesian Optimization (TuRBO1) \citep{erikssonScalableGlobalOptimization2020}, and the \verb|BoTorch| implementation of Vanilla BO \citep{balandat2020botorch}. On BayesMark, we also compare to LLAMBO \citep{liuLargeLanguageModels2024a}; for this method, we did not re-run experiments but instead relied on the results provided in the authors’ public GitHub repository \citep{liuTennisonliuLLAMBO2025}. As a consequence, we limit our empirical analysis to a constant total budget of $30$ function calls, regardless of the dimensionality of the problem.
Further details about the experimental procedures can be found in Appendix \ref{appendix:exp_details}.

\section{Results and Discussion}
\label{results}

\begin{figure}[ht]
     \centering
     \begin{subfigure}[b]{0.3\textwidth}
         \centering
         \includegraphics[width=\textwidth]{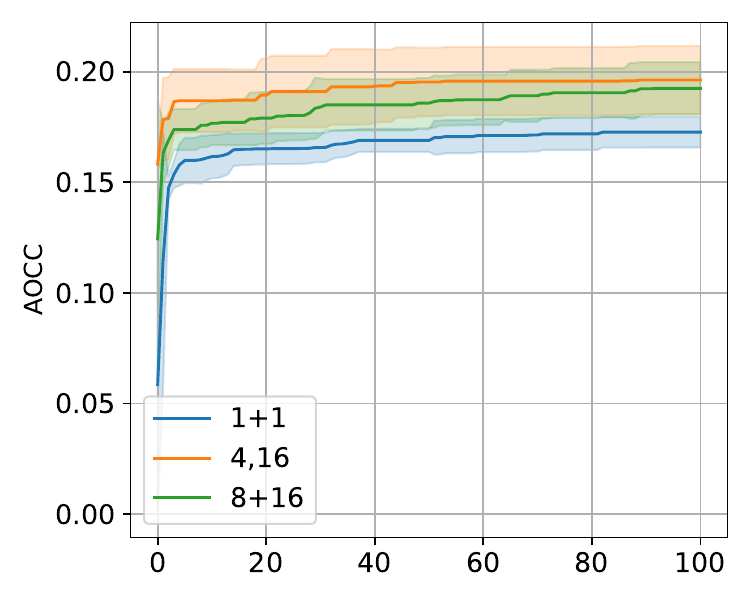}
         \caption{AOCC for different ES configurations.}
         \label{fig:BBOB_AOCC}
     \end{subfigure}
     \hfill
     \begin{subfigure}[b]{0.69\textwidth}
         \centering
        \includegraphics[width=\textwidth]{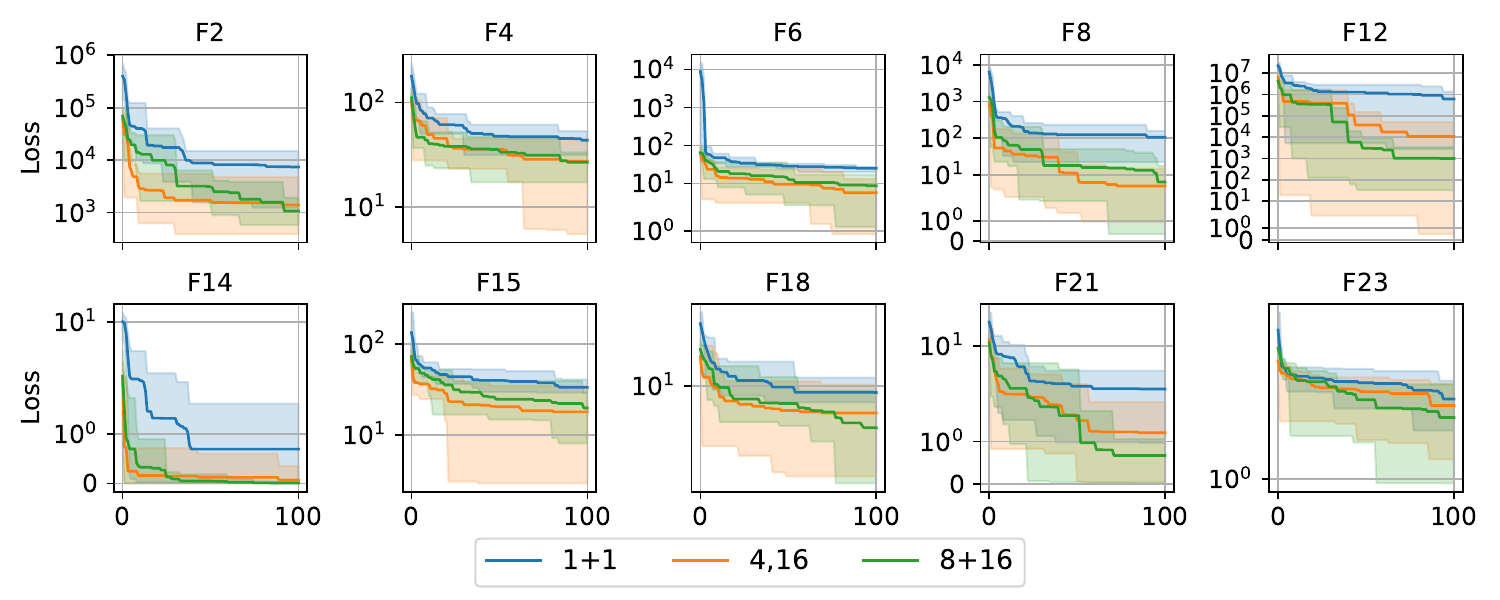}
         \caption{Loss of best generated algorithm at each evaluation of LLaMEA-BO.}
         \label{fig:BBOB_loss}
     \end{subfigure}
        \caption{LLaMEA-BO's performance on selected BBOB functions in terms of AOCC and generated algorithm loss over time. Shaded areas denote the standard error.}
        \label{fig:performance_LLM_evol}
\end{figure}

\textbf{Analysis of LLaMEA-BO performance.} In Figure \ref{fig:performance_LLM_evol}, we evaluate  the effectiveness of LLaMEA-BO using different evolution strategy (ES) configurations, focusing on two metrics: aggregate optimization performance via AOCC (as defined in Equation \eqref{eq:AOCC}) and per-function convergence behavior. Figure \ref{fig:BBOB_AOCC} reports the AOCC for three ES variants, aggregating results over a representative subset of 10 BBOB functions: the baseline $(1+1)$ strategy, and two population-based configurations $(4,16)$ and $(8+16)$. We observe that increasing population size improves AOCC, indicating that larger populations and a \textit{comma strategy} enable more effective exploration of the algorithm design space due to the less conservative evolutionary approach.
Figure \ref{fig:BBOB_loss} tracks the loss of the best generated algorithm over time on the single functions. Consistently with Figure  \ref{fig:BBOB_AOCC}, the ES configurations with highest population sizes, especially the $(4,16)$ strategy, 
lead to more rapid and consistent reductions in loss compared to the $(1+1)$ baseline.
In particular, the $(1+1)$ strategy consistently underperforms compared to the other configurations. This is primarily due to its limited ability to recover from poor generations: when an invalid algorithm is proposed, it receives an AOCC score of zero, and with no population diversity or crossover to guide recovery, such failures often stall progress. In contrast, larger ES configurations can simultaneously explore multiple candidates, making it more likely that at least one viable algorithm emerges and propagates over the search.

\begin{figure}[ht] \center
    \includegraphics[width=\columnwidth]{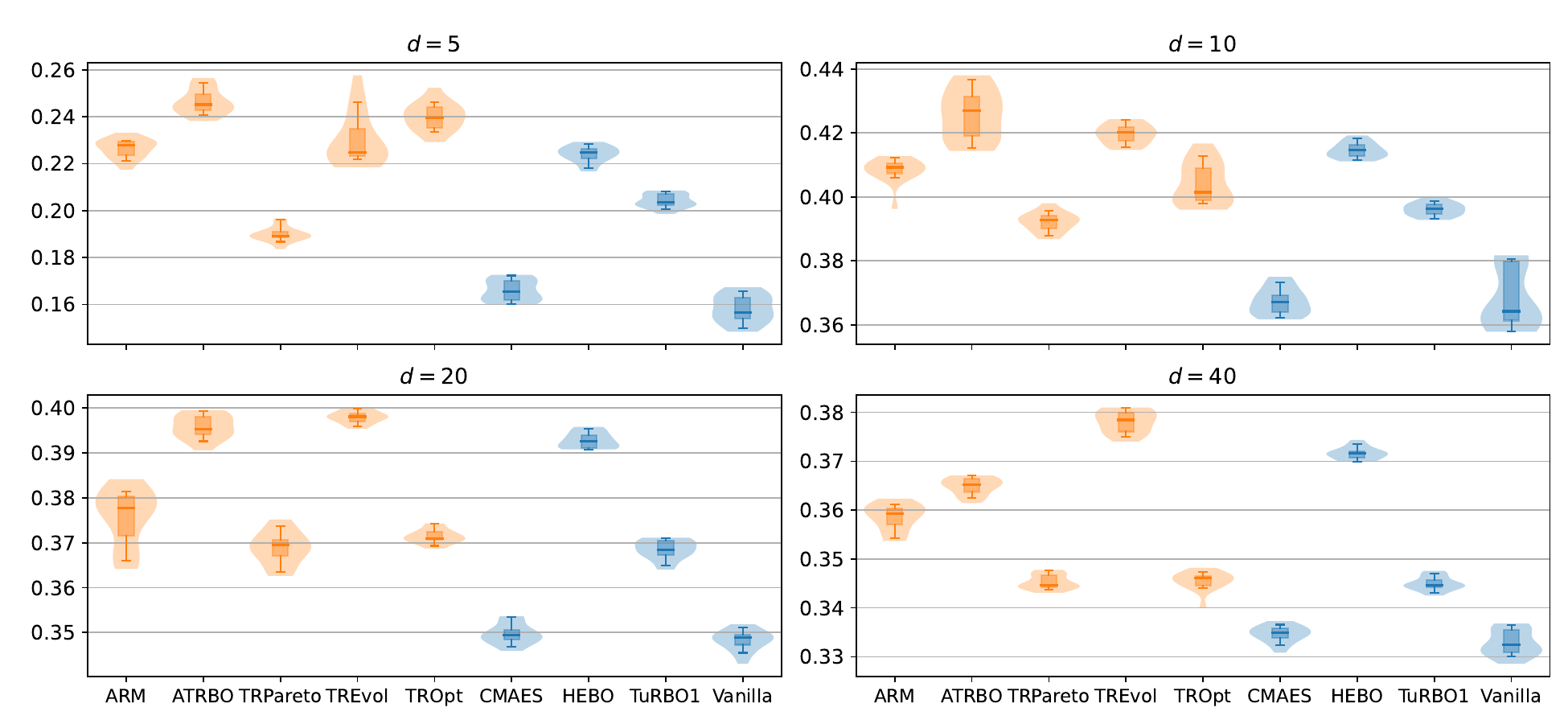}
      \caption{Best algorithm evaluation based on AOCC: Violin plots aggregating over 24 functions, 3 instances, 5 runs.}
    \label{fig:all_aoc_exc}
\end{figure}

\textbf{Evaluation of best generated algorithms on full BBOB.} To validate our proposed framework, we assess the extent to which the best generated algorithms generalize beyond the evaluations conducted during the optimization loop.
Figure \ref{fig:all_aoc_exc} presents the performance of the best BO algorithms generated by LLaMEA-BO in terms of average AOCC, evaluated on the full BBOB suite and aggregated over 24 functions, 3 instances, and 5 independent runs per setting. Each subplot corresponds to a different problem dimensionality $d \in \{5, 10, 20, 40\}$, and compares LLaMEA-generated algorithms (orange) to the established state-of-the-art baselines CMA-ES, HEBO, TuRBO1, and VanillaBO (blue). 
Across all dimensions, LLaMEA-BO generates several algorithm configurations that match or exceed the performance of state-of-the-art baselines. Among the generated algorithms, ATRBO performs best in lower dimensions (5 and 10), while TREvol leads in higher dimensions (20 and 40). Interestingly, both algorithms incorporate trust-region mechanisms (pseudocode is provided in Appendix \ref{app:ATRBO}). The consistent superiority of these methods across dimensions, as further supported by the results in Table \ref{table:aoc_results}, indicates strong generalization capabilities and suggests that the algorithms do not overfit to the specific problem settings used in the generation phase.
In particular, we notice that, despite being second-best during the code generation phase in $d=5$, TREvol consistently outperforms all compared state-of-the-art baselines (CMA-ES, HEBO, TuRBO1, and VanillaBO) across all dimensions. Complete convergence curves for each dimension are provided in Appendix~\ref{App:full_results_BBOB}.

\begin{table}
\caption{Average Area Over Convergence Curve over all $24$ BBOB functions, $3$ instances and $5$ random seeds. \textbf{Boldface} indicates the best value per dimension setting, \underline{underlined} text indicated statistical significance using a Paired T-test with $\alpha = 0.05$ of the best generated BO algorithm versus the best BO baseline. Algorithms are split in two groups: the ones generated by LLaMEA-BO on the left, the state-of-the-art baselines on the right.}
\label{table:aoc_results}
\centering
\resizebox{\textwidth}{!}{%
\begin{tabular}{@{}c|ccccc|cccc@{}}
\toprule
Dim & ARM & ATRBO & TREvol & TROpt & TRPareto & CMAES & HEBO & TuRBO1 & Vanilla \\
\midrule
5 & 0.2264 & \underline{\textbf{0.2464}} & 0.2304 & 0.2401 & 0.1899 & 0.1659 & 0.2241 & 0.2041 & 0.1581 \\
10 & 0.4083 & \underline{\textbf{0.4257}} & 0.4197 & 0.4043 & 0.3922 & 0.3673 & 0.4146 & 0.3961 & 0.3680 \\
20 & 0.3758 & 0.3958 & \textbf{0.3980} & 0.3714 & 0.3688 & 0.3497 & 0.3928 & 0.3685 & 0.3483 \\
40 & 0.3586 & 0.3649 & \textbf{0.3781} & 0.3455 & 0.3453 & 0.3346 & 0.3716 & 0.3450 & 0.3328 \\
\bottomrule
\end{tabular}}
\end{table}

\begin{figure}[ht] \center
    \includegraphics[width=\columnwidth]{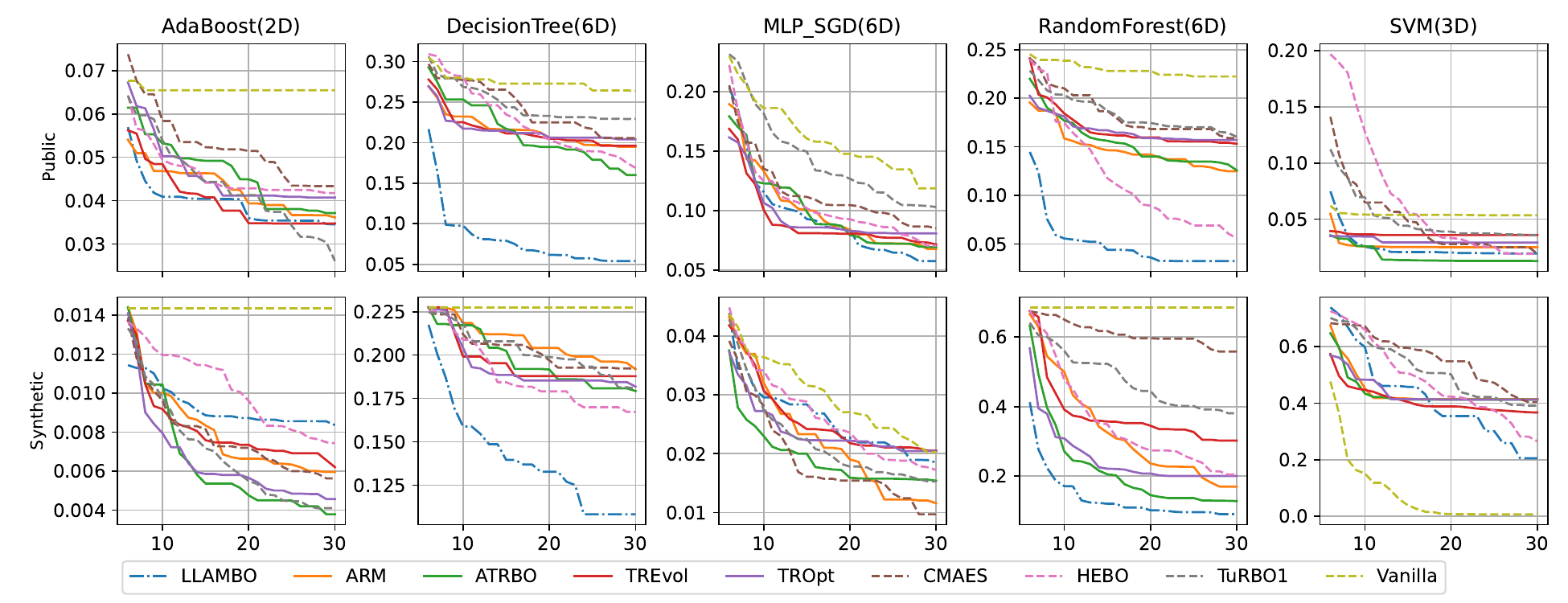}
      \caption{Hyperparameter tuning task results. Performance is reported in terms of regret and separately for Bayesmark public datasets (top row) and synthetic tasks (bottom row). Convergence curves are aggregated over $5$ independent runs. Solid lines correspond to our generated algorithms. All algorithms are initialized from the same $5$ samples, whose regret is not reported in the plots.}
    \label{fig:model_regret}
\end{figure}

\textbf{Robustness of the algorithms beyond BBOB: Analysis on Bayesmark.}  
In line with the study by \cite{liuLargeLanguageModels2024a}, Figure \ref{fig:model_regret} shows the performance on a set of unseen hyperparameter tuning tasks (averaged across different datasets) on both public datasets from the Bayesmark test suite, and synthetic datasets, presented in Section \ref{sec:exp_set_validation}. 
We observe that our direct competitor LLAMBO achieves a significantly better performance than our generated baselines in the DecisionTree and RandomForest tasks, both on public and synthetic datasets.
This is likely due to memorization of generally good hyperparameters for widely established and used ML models and benchmarks. 
LLaMEA-BO's algorithms demonstrate competitive performance on the other hyperparameter tuning (HPT) tasks, especially for models with smoother loss landscapes (the tree-based learners have mostly discrete hyper-parameters). 
Complete convergence curves for each HPT task are provided in Appendix~\ref{App:full_results_Bayesmark}.
These results confirm the robustness of the generated algorithms and suggest that LLaMEA-BO is capable of producing solutions that transfer effectively across domains without retraining or tuning. Convergence plots for the single task (model-dataset pair) are provided and discussed in Appendix \ref{App:full_results}.

\section{Conclusions}
\label{sec:conclusions}
In this paper, we introduced LLaMEA-BO, an extension of the LLaMEA framework designed for the automatic generation of Bayesian optimization (BO) algorithms by evolving full algorithm code through LLM in-context prompting and performance-driven feedback. Specifically, the LLMs are guided by an evolution strategy with in-loop evaluation on a representative set of functions and instances from the BBOB benchmarking suite within the COCO environment.

Experiments on the BBOB suite show that the best LLaMEA-BO-generated algorithms consistently match or outperform state-of-the-art BO baselines across various functions, dimensions, and instances. These algorithms also generalize effectively to unseen settings and maintain strong performance on hyperparameter tuning tasks from the Bayesmark benchmark.

Despite the goal of our research is not to generate a specific BO algorithm with absolute transversal supremacy across different testing settings, LLaMEA-BO proved to be a robust algorithm generator for continuous optimization tasks, highlighting the potential of LLMs not only as code synthesizers but as powerful co-designers of BO optimization algorithms. 

\textbf{Limitations.} 
While LLaMEA--BO achieves state-of-the-art performance on both BBOB and Bayesmark, several caveats remain.  
(i) Most high-performing discoveries are trust-region (\textsc{TR}) variants, hinting at an inductive bias introduced by the BO template and the short evaluation budget; other BO paradigms (e.g.\ information-theoretic or batch policies) are thus under-represented.  
(ii) All experiments rely on a single LLM backend (\texttt{gemini-2.0-flash}). Due to budget constraints experimenting with other LLMs was out of scope for this work. 
(iii) Because LLM sampling is inherently stochastic and downstream libraries evolve, exact reproducibility cannot be guaranteed; repeated runs may generate slightly different algorithms. To partly mitigate this limitation we open-source all raw logs and LLM queries from each experiment. 
(iv) The empirical study is limited to noiseless, continuous, single-objective tasks; mixed-integer, constrained, noisy, or multi-objective settings are left unexplored.

\textbf{Outlook. }
These limitations suggest several natural extensions.  
(i) Diversifying prompts, variation operators, and fitness aggregation should encourage structural innovations beyond \textsc{TR} logic, while querying a portfolio of LLMs could enhance robustness.  
(ii) Because the generate–evaluate–improve loop is benchmark-agnostic, integrating mixed-integer, multi-objective, constrained, or noise-aware benchmarks requires only minor modifications.

\clearpage



\small
\bibliographystyle{abbrvnat}
\bibliography{llameabo_arxiv_clean.bib}







\newpage

\appendix

\section{Further Experimental Details}
\label{appendix:exp_details}

This section documents additional details on the evaluation pipeline: Subsection \ref{appendix:metrics}
formalises the \textit{AOCC} score and its aggregation, while
Subsection \ref{sec:safe_execution} describes the sandbox that prevents untrusted code from hanging or leaking resources.

\subsection{AOCC Performance Metric Definition}
\label{appendix:metrics}

\textit{AOCC} measures the area over the convergence curve, which is equivalent to the area under the log-scaled precision CDF,
rewarding early progress yet capping pathological values via
tight \textit{lb}/\textit{ub} bounds. Averaging over functions,
instances and repeats yields a scalar
fitness that is smooth and makes ranking between different algorithms based on a complete benchmark set possible.

\paragraph{AOCC definition.}
Let $\mathbf{y}_{a,f}$ be the sequence of \emph{best‑seen} objective values
obtained by algorithm~$a$ on function instance~$f$ after each evaluation.
Following~\citep{lopez2014automatically},
we log‑scale the precision values and normalize them
between the lower and upper bounds (we use
$\textit{lb} = 10^{-8}$ and $\textit{ub} = 10^{4}$ for BBOB $5d$ and an increased upper bound of $10^{9}$ for $10, 20, \text{ and }  40d$):
\begin{equation}
  \label{eq:AOCC}
  \textit{AOCC}(\mathbf{y}_{a,f})
  = \frac{1}{B}\sum_{i=1}^{B}
    \Bigl(
      1 -
      \frac{
        \min\bigl(\max(y_i,\textit{lb}),\textit{ub}\bigr) - \textit{lb}
      }{
        \textit{ub} - \textit{lb}
      }
    \Bigr)\;,
\end{equation}
where $y_i = \log\!\bigl(f(x_i) - f^\star\bigr)$ is the
log‑precision after the $i$‑th evaluation
and $f^\star$ is the global optimum of $f$.
Equation~\eqref{eq:AOCC} can be interpreted as
the area under the empirical cumulative distribution function (CDF) over an infinite grid of target precisions.

\paragraph{Aggregation across functions, instances and runs.}
For each algorithm~$a$ we compute the mean $AOCC$ over the
$N_f$ functions and $N_j$ instances per function:
\begin{equation}
  \textit{AOCC}(a)
  = \frac{1}{N_i \cdot N_f}\;
    \sum_{j=1}^{N_f}\sum_{k=1}^{N_i}
    \textit{AOCC}\ \bigl(\mathbf{y}_{a,f_{jk}}\bigr)\;.
\end{equation}
We then average over $N_s$ independent repetitions
to obtain the scalar fitness feedback to the LLM:
\begin{equation}
  \label{eq:fitness}
  f(a)
  = \frac{1}{N_s}\sum_{s=1}^{N_s}\textit{AOCC}^{(s)}(a)\;.
\end{equation}
Any runtime, compilation or import error during evaluation
is assigned an $AOCC$ of zero, propagating a minimum fitness to the
evolutionary loop.

\subsection{Test Problems}
\label{app:testbed}

We evaluate all algorithms on three distinct categories of optimization problems: standard noiseless functions from the BBOB suite from the COCO environment, machine learning hyperparameter tuning (HPT) tasks from Bayesmark, and synthetic benchmarks to assess generalization capabilities.

\subsubsection{BBOB}
\label{app:bbob}

The BBOB (Black-Box Optimization Benchmarking) suite~\citep{hansen2021coco} consists of 24 noiseless functions commonly used for evaluating optimization algorithms. These functions are grouped into five categories based on their characteristics: separable functions, functions with low or moderate conditioning, functions with high conditioning and unimodal landscape, multimodal functions with adequate global structure, and multimodal functions with weak global structure. We use all 24 functions across three random instances $(4,5,6)$---instances $(1,2,3)$ were used for the code generation phase with LLaMEA-BO---and dimensions \(d \in \{5, 10, 20, 40\}\).
For different instances, functions are shifted and rotated to ensure invariance to translation and scaling, and evaluated within the default search domain \([-5, 5]^d\). We applied no rescaling or normalization to the input space beyond the BBOB specifications.
Performance is assessed via the AOCC and loss metrics, averaged across instances and random seeds.

\subsubsection{Bayesmark}
\label{app:bayesmark}

As in \citep{liuLargeLanguageModels2024a}, we utilize Bayesmark~\citep{UberBayesmark2025} as a continuous HPT benchmark. We include the 5 public datasets provided in the benchmark and 5 ML models: RandomForest, SVM, DecisionTree, MLP, and AdaBoost, resulting in 25 unique (dataset, model) tasks. Each task is executed using $5$ different random seeds, with consistent model initialization across all runs to ensure fair comparisons. Classification tasks are evaluated using accuracy, while regression tasks use MSE as the scoring metric.

\paragraph{Hyperparameter space.} We adopt the search spaces and ranges defined in Bayesmark for each model, including parameter types (e.g., linear, log, logit) and bounds. The spaces used for each model are summarized as follows:

\begin{itemize}
    \item \textbf{SVM (3d):} 
    \texttt{C} [log; 1, 1e3], 
    \texttt{gamma} [log; 1e-4, 1e-3], 
    \texttt{tolerance} [log; 1e-5, 1e-1]
    
    \item \textbf{DecisionTree (6d):} 
    \texttt{max\_depth} [linear; 1, 15], 
    \texttt{min\_samples\_split} [logit; 0.01, 0.99], 
    \texttt{min\_samples\_leaf} [logit; 0.01, 0.49], 
    \texttt{min\_weight\_fraction\_leaf} [logit; 0.01, 0.49], 
    \texttt{max\_features} [logit; 0.01, 0.99], 
    \texttt{min\_impurity\_decrease} [linear; 0.0, 0.5]

    \item \textbf{RandomForest (6d):} same as DecisionTree

    \item \textbf{MLP (6d):} 
    \texttt{hidden\_layer\_sizes} [linear; 50, 200], 
    \texttt{alpha} [log; 1e-5, 1e1], 
    \texttt{batch\_size} [linear; 10, 250], 
    \texttt{learning\_rate\_init} [log; 1e-5, 1e-1], 
    \texttt{power\_t} [logit; 0.1, 0.9], 
    \texttt{momentum} [logit; 0.001, 0.999] 

    \item \textbf{AdaBoost (2d):} 
    \texttt{n\_estimators} [linear; 10, 100], 
    \texttt{learning\_rate} [log; 1e-4, 1e1]
\end{itemize}

\subsubsection{Synthetic Benchmarks}
\label{app:synthetic}

In addition to public data, we evaluate algorithms on 3 synthetic datasets designed to assess generalization, in line with the analysis by \cite{liuLargeLanguageModels2024a}. The synthetic benchmarks are generated from multimodal functions (Rosenbrock, Griewank, and KTablet) with input dimensionality set to 15. Each input is sampled uniformly from \([0, 1]\), and the corresponding output is computed using the respective analytical function~\citep{watanabeTreeStructuredParzenEstimator2023b}. These datasets are integrated into Bayesmark, evaluated with the same machine learning models and protocols described in Appendix~\ref{app:bayesmark}, yielding 15 tasks in total. 

\subsection{Safe Execution of Generated Algorithms}
\label{sec:safe_execution}

During evaluation of the generated algorithms, the generated code is loaded and executed inside a safe local environment. The generated algorithm has access to a curated list of popular Python packages used in BO such as \textit{scikit-learn} and \textit{pytorch}.
Each algorithm is first evaluated on a dummy problem to detect early runtime and syntax errors before the complete benchmarking is started. To prevent infinite runs, we limit each algorithm to a $30$ minutes execution time window for the complete benchmark evaluation.

\subsection{Broader Impact}
\label{appendix:broader-impact}

\paragraph{Positive impacts.}
By automating the design of BO algorithms,
LLaMEA-BO can shorten the development cycle in fields where
manual tuning is common but expert time is scarce, such as
computational biology, materials discovery and robotics.  
Because the generated code is lightweight and written in plain Python,
non-specialists can inspect, adapt and extend the resulting algorithms,
lowering barriers to entry and promoting methodological transparency.
All artefacts are released under a permissive licence, facilitating
open-science reuse and replication.

\paragraph{Potential risks.}
\emph{(i) Model misuse.}  
Although the generated code is benchmark-agnostic, an actor could tailor it for black-box attacks (e.g.\ model
inversion or adversarial example search).  
\emph{(ii) Compute footprint.}  
Large-scale prompting incurs non-negligible energy cost: a
single 100-generation search consumes roughly $20$ CPU-hours.

\paragraph{Mitigation measures.}
We sandbox all generated code (\S\ref{sec:safe_execution}) to mitigate unintended side effects.
Release artefacts include usage guidelines that discourage employing
the system for privacy-violating optimisation.  
Researchers are urged to rerun the search phase on tasks that match
their real-world noise, fidelity and constraint profiles, and to audit
solutions with domain-specific tests before deployment.

\subsection{Hardware and Runtime}
All experiments are conducted on a Linux cluster with 68 nodes with dual 8-core 2.4 GHz CPUs, 64 GB RAM per node, and a total of 128 TB shared storage. The compute time for the evolutionary loop highly depends on the generated code at each iteration, spanning from $2$ to $20$h wall-clock time for one run.

\section{Code Template}
\label{sec:code-template}

Listing~\ref{listing:template} shows the 
LLaMEA-BO boiler code that every LLM prompt inherits.
The template enforces a common API (\texttt{\_\_init\_\_}/\texttt{\_\_call\_\_})
and imports only whitelisted libraries, ensuring compatibility with popular BO frameworks.

%


\begin{table}[H]
	\centering
	\caption{Code template used in the initialisation prompt of LLaMEA-BO.}
	\label{listing:template}
\begin{lstlisting}[style=mintedstyle]
from collections.abc import Callable
from scipy.stats import qmc #If you are using QMC sampling.
from scipy.stats import norm
import numpy as np
class <AlgorithmName>:
    def __init__(self, budget:int, dim:int):
        self.budget = budget
        self.dim = dim
        # bounds has shape (2,<dimension>), bounds[0]: lower bound, bounds[1]: upper bound
        self.bounds = np.array([[-5.0]*dim, [5.0]*dim])
        # X has shape (n_points, n_dims), y has shape (n_points, 1)
        self.X: np.ndarray = None
        self.y: np.ndarray = None
        self.n_evals = 0 # the number of function evaluations
        self.n_init = <your_strategy>

        # Do not add any other arguments without a default value

    def _sample_points(self, n_points):
        # sample points
        # return array of shape (n_points, n_dims)

    def _fit_model(self, X, y):
        # Fit and tune surrogate model 
        # return the model
        # Do not change the function signature

    def _acquisition_function(self, X):
        # Implement acquisition function 
        # calculate the acquisition function value for each point in X
        # return array of shape (n_points, 1)

    def _select_next_points(self, batch_size):
        # Select the next points to evaluate
        # Use a selection strategy to optimize/leverage the acquisition function 
        # The selection strategy can be any heuristic/evolutionary/mathematical/hybrid methods.
        # Your decision should consider the problem characteristics, acquisition function, and the computational efficiency.
        # return array of shape (batch_size, n_dims)

    def _evaluate_points(self, func, X):
        # Evaluate the points in X
        # func: takes array of shape (n_dims,) and returns np.float64.
        # return array of shape (n_points, 1)

        self.n_evals += len(X)
    
    def _update_eval_points(self, new_X, new_y):
        # Update self.X and self.y
        # Do not change the function signature
    
    def __call__(self, func:Callable[[np.ndarray], np.float64]) -> tuple[np.float64, np.array]:
        # Main minimize optimization loop
        # func: takes array of shape (n_dims,) and returns np.float64. 
        # !!! Do not call func directly. Use _evaluate_points instead and be aware of the budget when calling it. !!!
        # Return a tuple (best_y, best_x)
        
        self._evaluate_points()
        self._update_eval_points()
        while self.n_evals < budget:
            # Optimization

            # select points by acquisition function
            self._evaluate_points()
            self._update_eval_points()

        return best_y, best_x
    
\end{lstlisting}

\end{table}

\section{Ablation Studies}
\label{appendix:ablation}

The following ablations stress-test the LLaMEA-BO algorithm.  All trends are
consistent across ten BBOB functions,
performance is resilient to most hyper-parameters, but a handful of settings  dominate.

\subsection{ES Configuration}

\begin{figure}[H]
     \centering
     \begin{subfigure}[b]{0.51\textwidth}
         \centering
         \includegraphics[width=\textwidth]{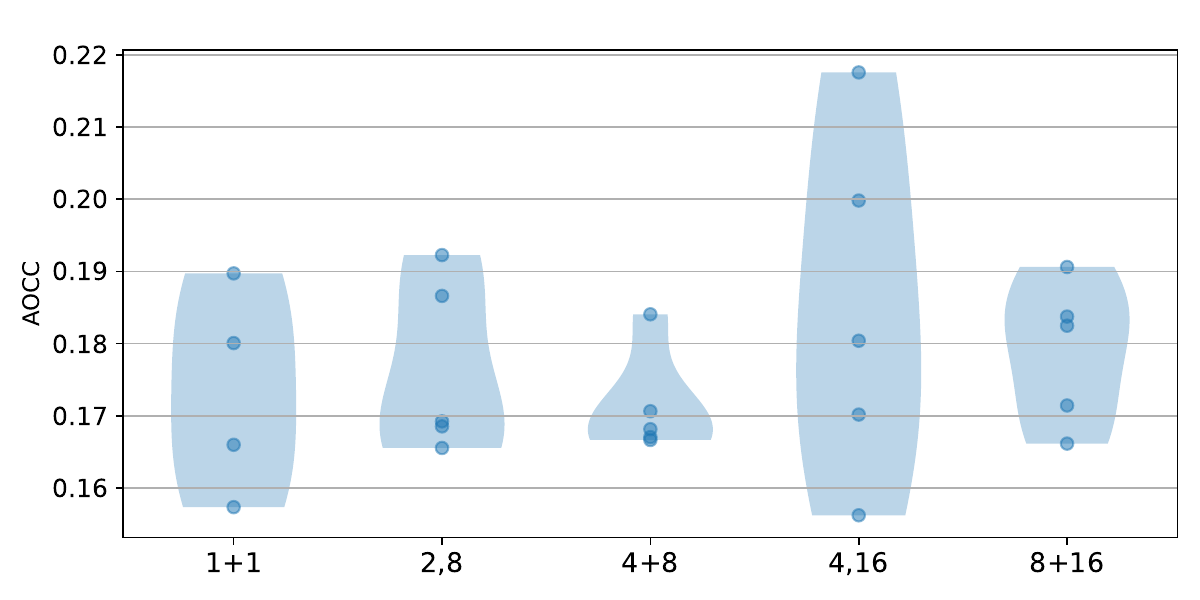}
         \caption{Final AOCC distributions of the different ES strategies.}
     \end{subfigure}
     \hfill
     \begin{subfigure}[b]{0.48\textwidth}
         \centering
        \includegraphics[width=\textwidth]{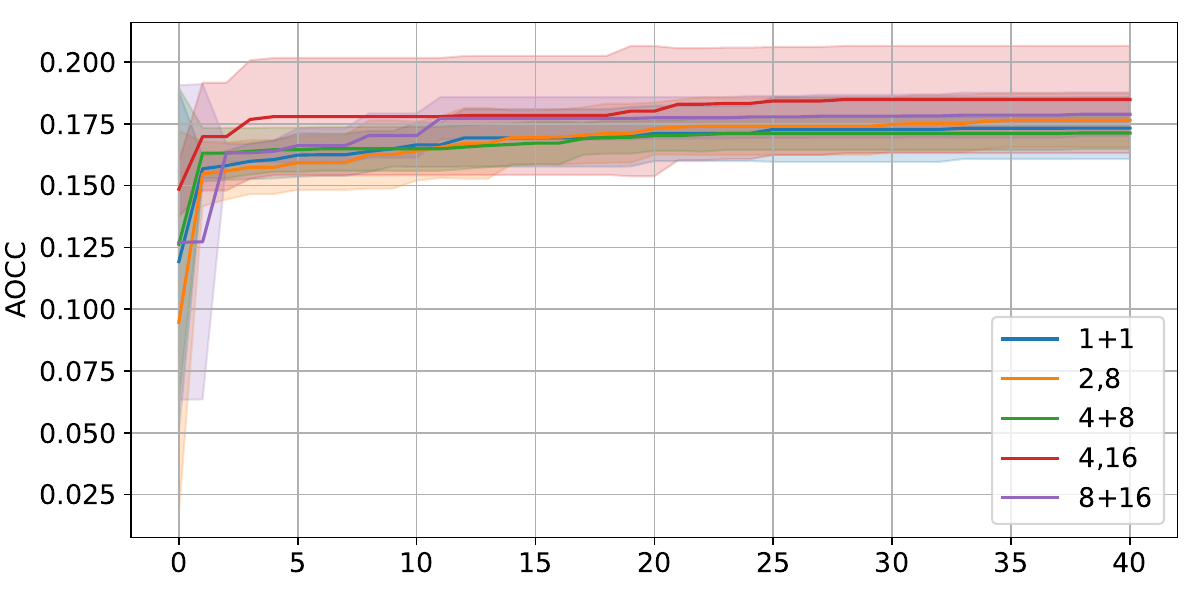}
         \caption{AOCC convergence curves of the different ES strategies.}
     \end{subfigure}
        \caption{Results from different population sizes and elitism configurations averaged over $10$ BBOB functions and $5$ repetitions per function.}
        \label{fig:ablation-es-strategies}
\end{figure}

\begin{figure}[H]
     \centering
     \begin{subfigure}[b]{0.51\textwidth}
         \centering
         \includegraphics[width=\textwidth]{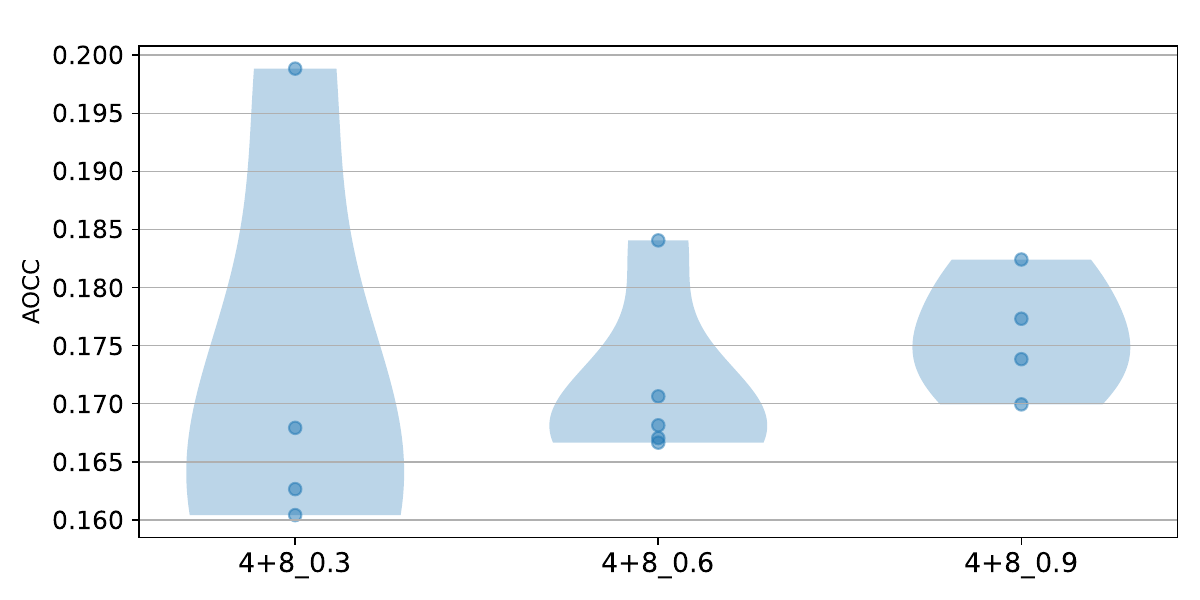}
         \caption{Final AOCC distributions.}
     \end{subfigure}
     \hfill
     \begin{subfigure}[b]{0.48\textwidth}
         \centering
        \includegraphics[width=\textwidth]{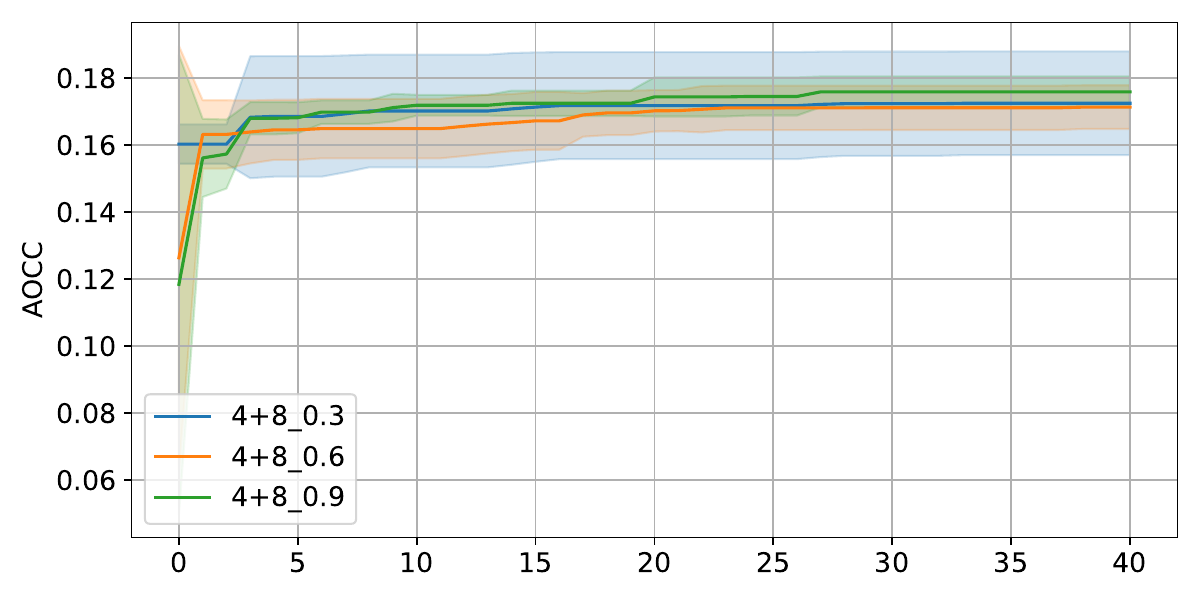}
         \caption{AOCC convergence curves.}
     \end{subfigure}
        \caption{Results from different crossover rate $[0.3, 0.6, 0.9]$ configurations (using a $(4+8)$ ES strategy) averaged over $10$ BBOB functions and $4$ repetitions per function.}
    \label{fig:ablation-es-crossover}
\end{figure}

Figure \ref{fig:ablation-es-strategies} shows that a non-elitist strategy gives the highest mean AOCC; larger populations improves the stability of the results across multiple runs.  As seen in Figure
\ref{fig:ablation-es-crossover}, a crossover rate of $0.6$ or $0.9$
achieves the best trade-off although not significantly.

\subsubsection{Population Size and Elitism}

\subsubsection{Crossover Rate}

\subsection{LLM Configuration}
\label{appendix:LLMconfig}

\begin{figure}[H]
     \centering
     \begin{subfigure}[b]{0.51\textwidth}
         \centering
         \includegraphics[width=\textwidth]{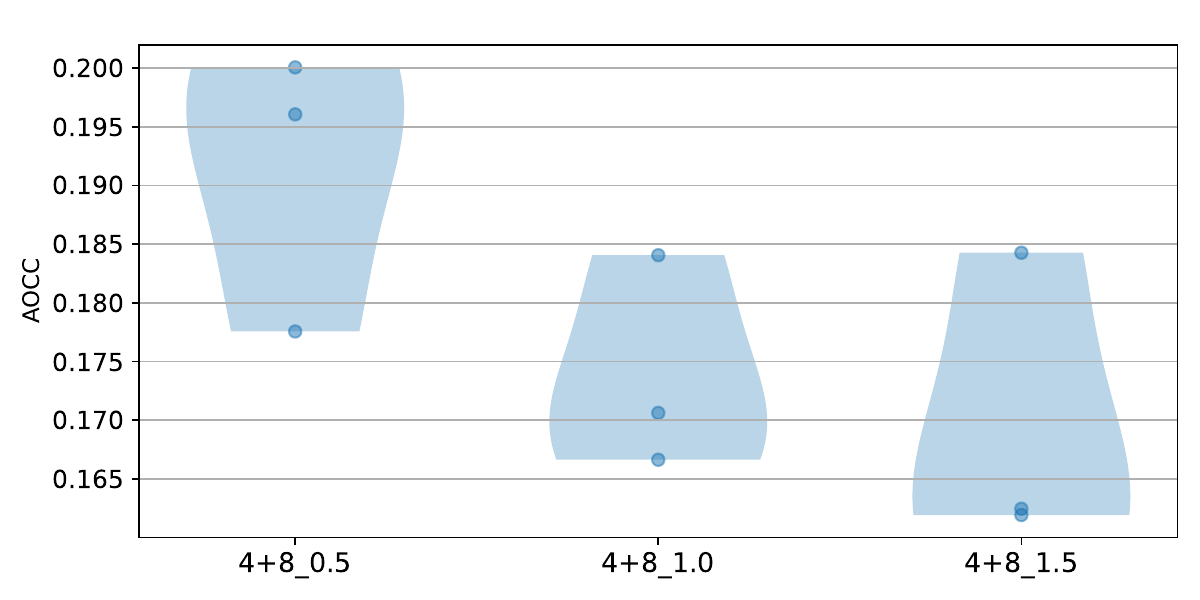}
         \caption{Final AOCC distributions.}
     \end{subfigure}
     \hfill
     \begin{subfigure}[b]{0.48\textwidth}
         \centering
        \includegraphics[width=\textwidth]{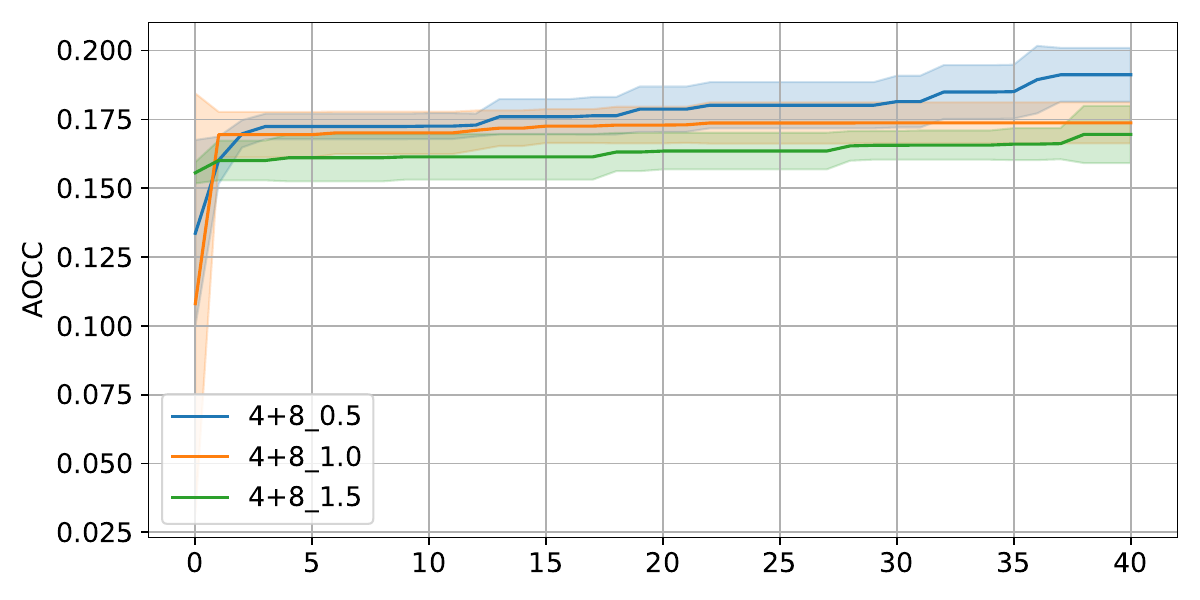}
         \caption{AOCC convergence curves.}
     \end{subfigure}
        \caption{Results from different LLM temperature $[0.5, 1.0, 1.5]$ configurations (using a $(4+8)$ ES strategy) averaged over $10$ BBOB functions and $3$ repetitions per function.}
    \label{fig:ablation-llm-temperature}
\end{figure}

\begin{figure}[H]
     \centering
     \begin{subfigure}[b]{0.51\textwidth}
         \centering
         \includegraphics[width=\textwidth]{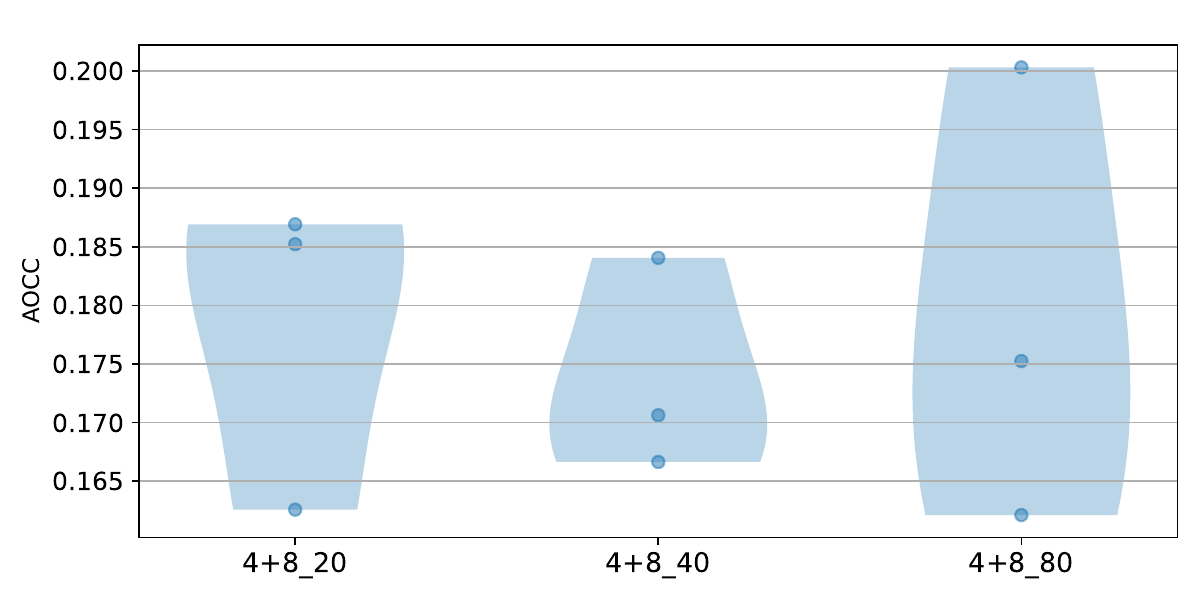}
         \caption{Final AOCC distributions.}
     \end{subfigure}
     \hfill
     \begin{subfigure}[b]{0.48\textwidth}
         \centering
        \includegraphics[width=\textwidth]{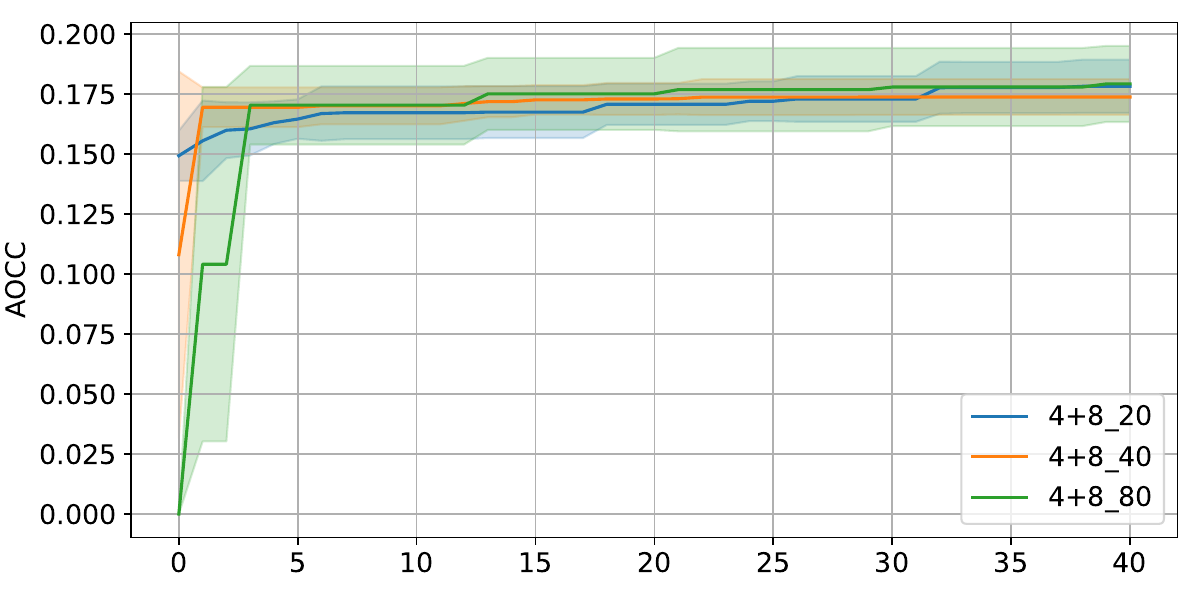}
         \caption{AOCC convergence curves.}
     \end{subfigure}
        \caption{Results from different LLM top-K $[20, 40, 80]$ configurations (using a $(4+8)$ ES strategy) averaged over $10$ BBOB functions and $3$ repetitions per function.}
    \label{fig:ablation-llm-topk}
\end{figure}

\begin{figure}[H]
     \centering
     \begin{subfigure}[b]{0.51\textwidth}
         \centering
         \includegraphics[width=\textwidth]{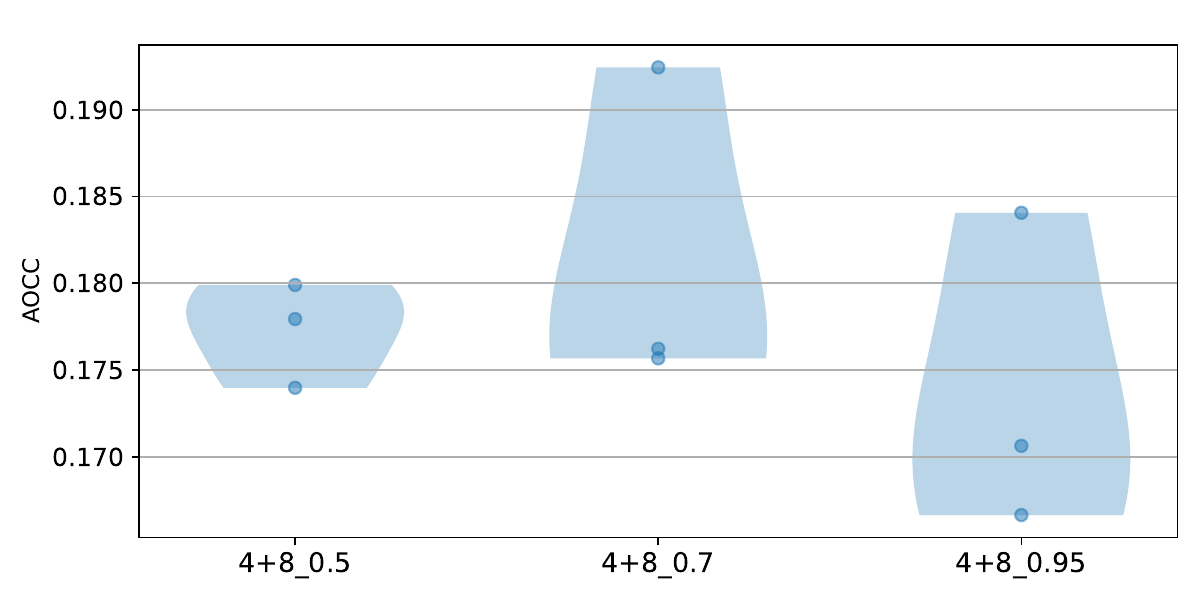}
         \caption{Final AOCC distributions.}
     \end{subfigure}
     \hfill
     \begin{subfigure}[b]{0.48\textwidth}
         \centering
        \includegraphics[width=\textwidth]{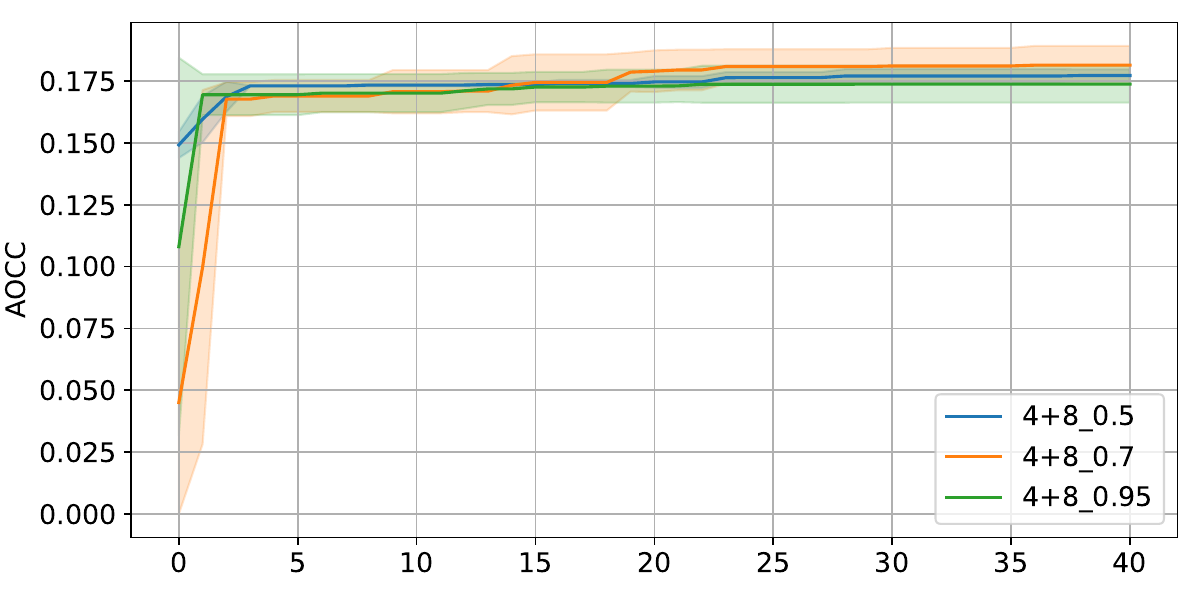}
         \caption{AOCC convergence curves.}
     \end{subfigure}
    \caption{Results from different LLM top-P $[0.5, 0.7, 0.95]$ configurations (using a $(4+8)$ ES strategy) averaged over $10$ BBOB functions and $3$ repetitions per function.}
    \label{fig:ablation-llm-topP}
\end{figure}

Figures \ref{fig:ablation-llm-temperature}, 
\ref{fig:ablation-llm-topk}, \ref{fig:ablation-llm-topP} reveal that sampling temperature
$0.5$, top-$k=40$ and top-$p=0.7$ form a sweet spot, more conservative
decoding reduces diversity, whereas aggressive sampling inflates
invalid code rates. Overall the LLM performance is quite stable and not very sensitive to these settings in our context.

\subsubsection{Temperature}

\subsubsection{Top-K}

\subsubsection{Top-P}

\section{Excluding Biases in Benchmarking}

\begin{figure}[H]
    \centering
    \includegraphics[width=.8\textwidth,trim=0mm 0mm 0mm 0mm,clip]{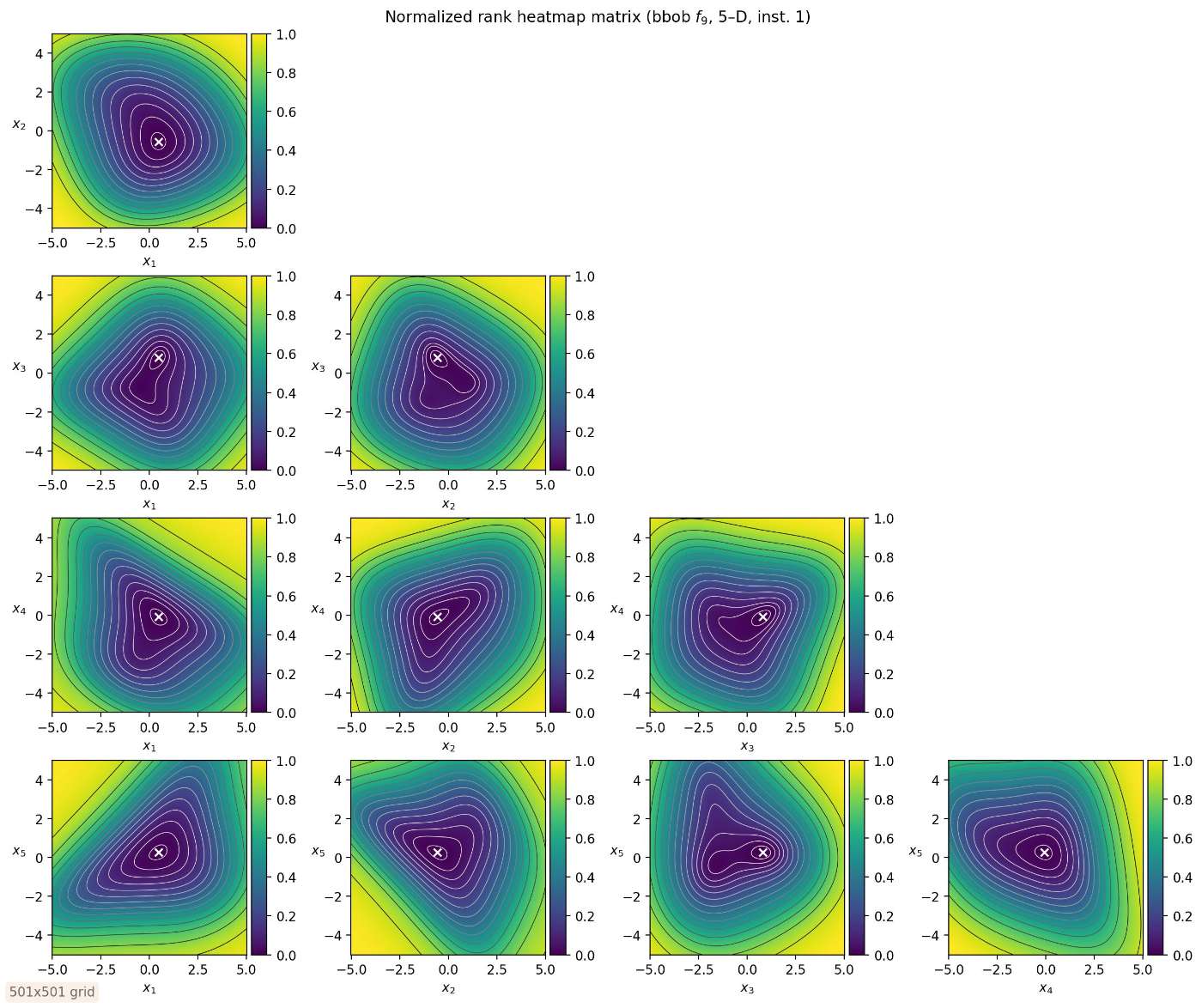}
    \caption{Level sets for BBOB function $f_9$ (instance $id=1$) in $d=5$, showing the optimum near the center of the search domain. Figure from \url{https://coco-platform.org/testsuites/bbob/functions/f09.html}.}
    \label{fig:bbob-bias-levelsets}
\end{figure}

\begin{figure}[H]
    \centering
    \includegraphics[width=\textwidth]{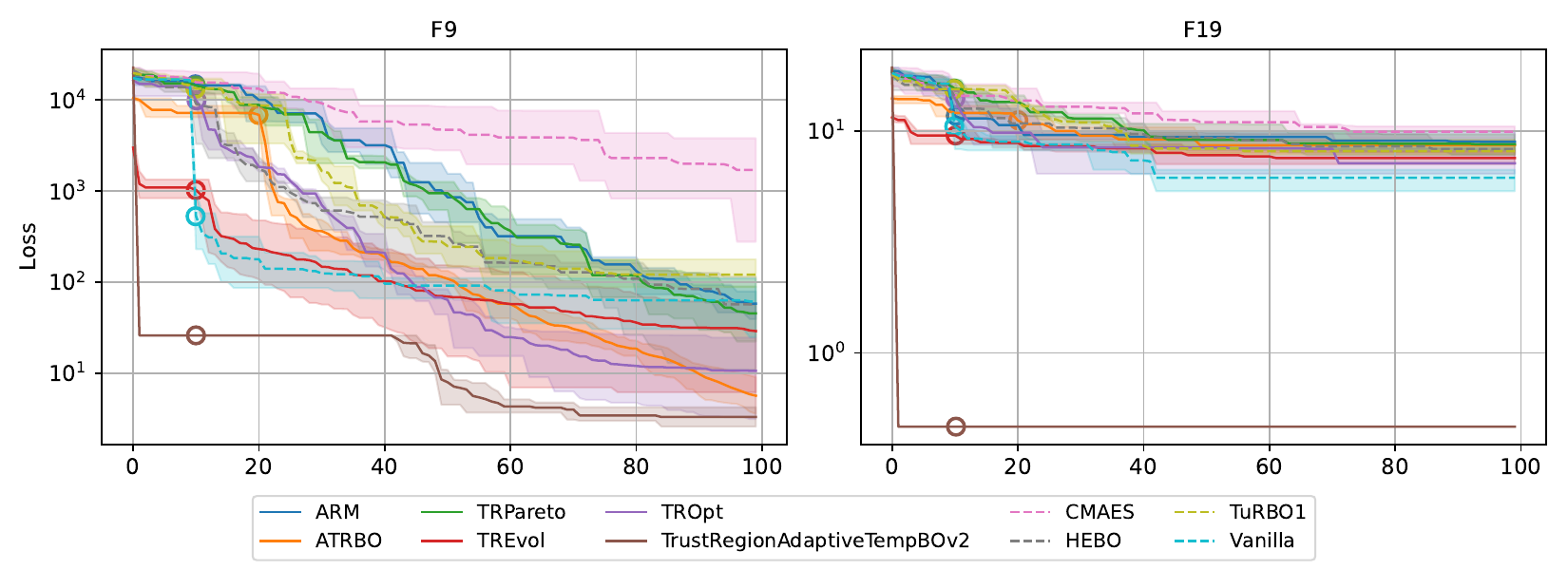}
    \caption{Observed strange behaviour of ``TrustRegionAdaptiveTempBOv2'' on $f_9$ and $f_{19}$}
    \label{fig:bbob-bias}
\end{figure}

Figures \ref{fig:bbob-bias-levelsets} illustrates how
the optimum of $f_9$ lies near the search-space origin, the same is known for $f_{19}$ \cite{long2023bbob}. These biased optima locations
allows one generated algorithm to basically ``cheat'' by sampling near the origin $(0,\dots,0)$. We can see that this actually occurs in Figure \ref{fig:bbob-bias}. Here we have one algorithm called ``TrustRegionAdaptiveTempBOv2" with very good performance already from the very first evaluation. This bias in sampling strategies does not generalize normally and is not preferred.
We therefore exclude those biased instances from the evaluation procedure of the final generated BO algorithms. Note that we did not have these functions included in the training procedure to begin with.

\section{Top-generated Algorithms}
\label{App: top5}

Table~\ref{table:llamea_algorithms} provides an overview of the top-5 algorithms automatically generated by LLaMEA-BO. For each algorithm, we report its short and full name, the population configuration used during its evolution (elitist $(\mu{+}\lambda)$ or non-elitist $(\mu,\lambda)$, its performance in terms of average area over the convergence curve (AOCC) across the 10 BBOB functions chosen in the evolutionary loop for algorithm generation, and the evolutionary operator responsible for its creation (initialization, crossover, or mutation). These algorithms were selected based on their overall AOCC performance and exhibit a diverse range of algorithmic characteristics---trust regions, enhanced exploration, Pareto fronts, ensembles, adaptive components---underscoring the capability of LLaMEA-BO to generate varied BO pipelines. However, the incorporation of trust regions appears to be a key contributor to strong performance.

\begin{table}[h!]
\caption{Generated algorithms from LLaMEA-BO. Each algorithm is labeled with its name, short description, population configuration ($\mu{+}\lambda$), AOCC(Search) score, and its origin within the evolutionary process (Initialization, Mutation, or Crossover).}
\label{table:llamea_algorithms}
\centering
\resizebox{\textwidth}{!}{%
\begin{tabular}{@{}lp{10cm}ccl@{}}
\toprule
\textbf{Algorithm} & \textbf{Name} & \textbf{ES Configuration} & \textbf{AOCC (Search)} & \textbf{Type} \\
\midrule
ATRBO & Adaptive Trust Region Bayesian Optimization & (4,16) & 0.2091 & Initialization \\
\addlinespace
TREvol & Adaptive Trust Region Evolutionary BO with Dynamic Kernel, Acquisition Blending, Adaptive DE, Gradient-Enhanced Trust Region Adjustment, and Variance-Aware Exploration (ATREBO-DKA-BDE-GE-VAE) & (8+16) & 0.2138 & Crossover \\
\addlinespace
TROpt & Adaptive Trust Region Optimistic Hybrid BO & (4,16) & 0.2043 & Crossover \\
\addlinespace
TRPareto & Adaptive Evolutionary Pareto Trust Region BO & (8+16) & 0.1827 & Mutation \\
\addlinespace
ARM & Adaptive Batch Ensemble with Thompson Sampling, Density-Aware Exploration, Uncertainty-Aware Local Search with Adaptive Radius and Momentum Bayesian Optimization (ABETSALSED\_ARM\_MBO) & (8+16) & 0.1813 & Crossover \\
\bottomrule
\end{tabular}}
\end{table}

\section{Full Convergence Results}
\label{App:full_results}

In this section we provide the complete per-function curves on BBOB functions (loss) and Bayesmark tasks (regret) for the resulting generated BO algorithms (bold lines) versus BO baselines (dashed lines).

\subsection{BBOB}
\label{App:full_results_BBOB}
\begin{figure}[H]
    \centering
    \includegraphics[width=\textwidth,trim=0mm 0mm 0mm 0mm,clip]{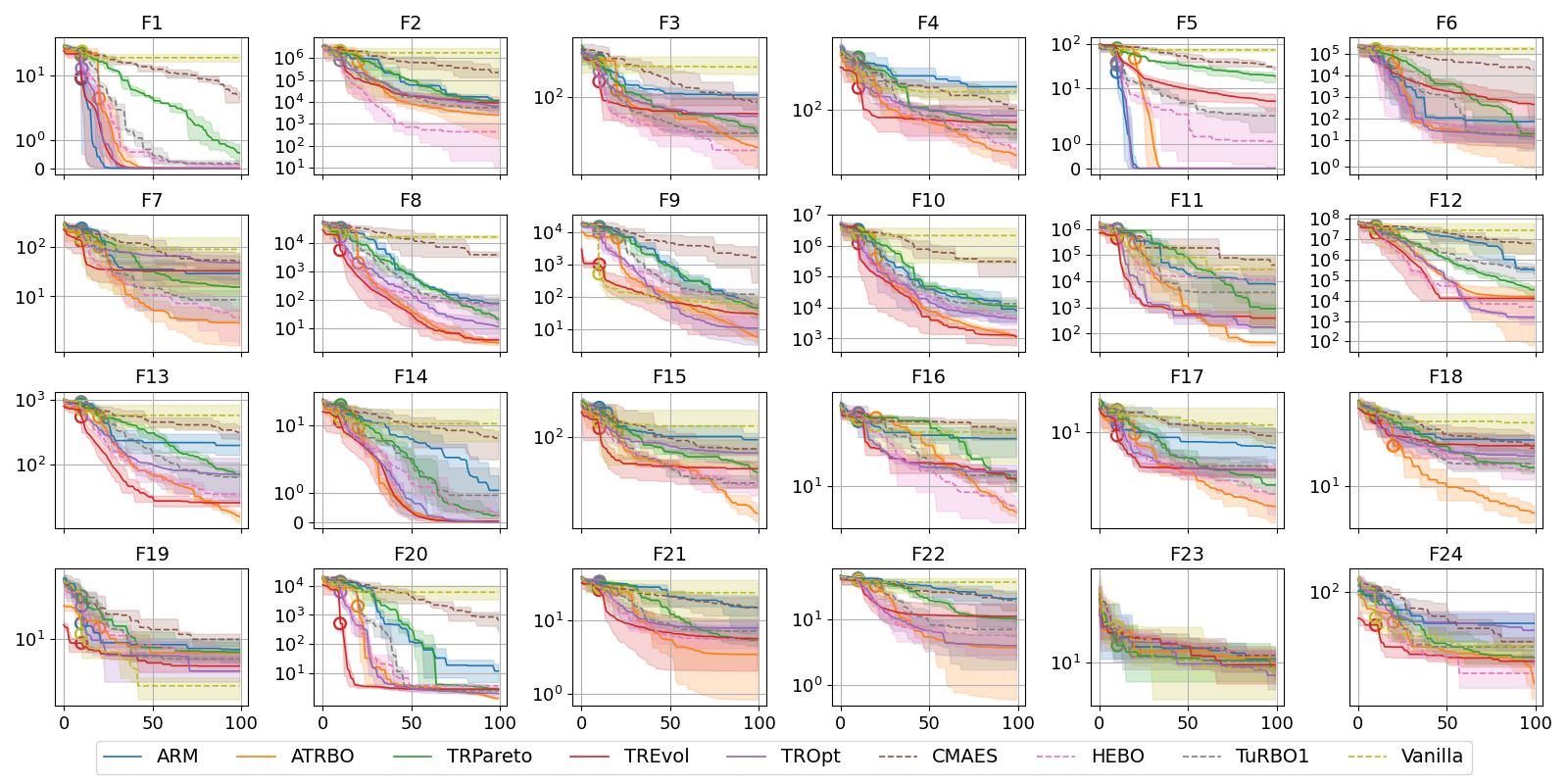}
    \caption{Loss convergence curves on all  BBOB functions in $d=5$.}
    \label{fig:bbob-5d}
\end{figure}

\begin{figure}[H]
    \centering
    \includegraphics[width=\textwidth,trim=0mm 0mm 0mm 0mm,clip]{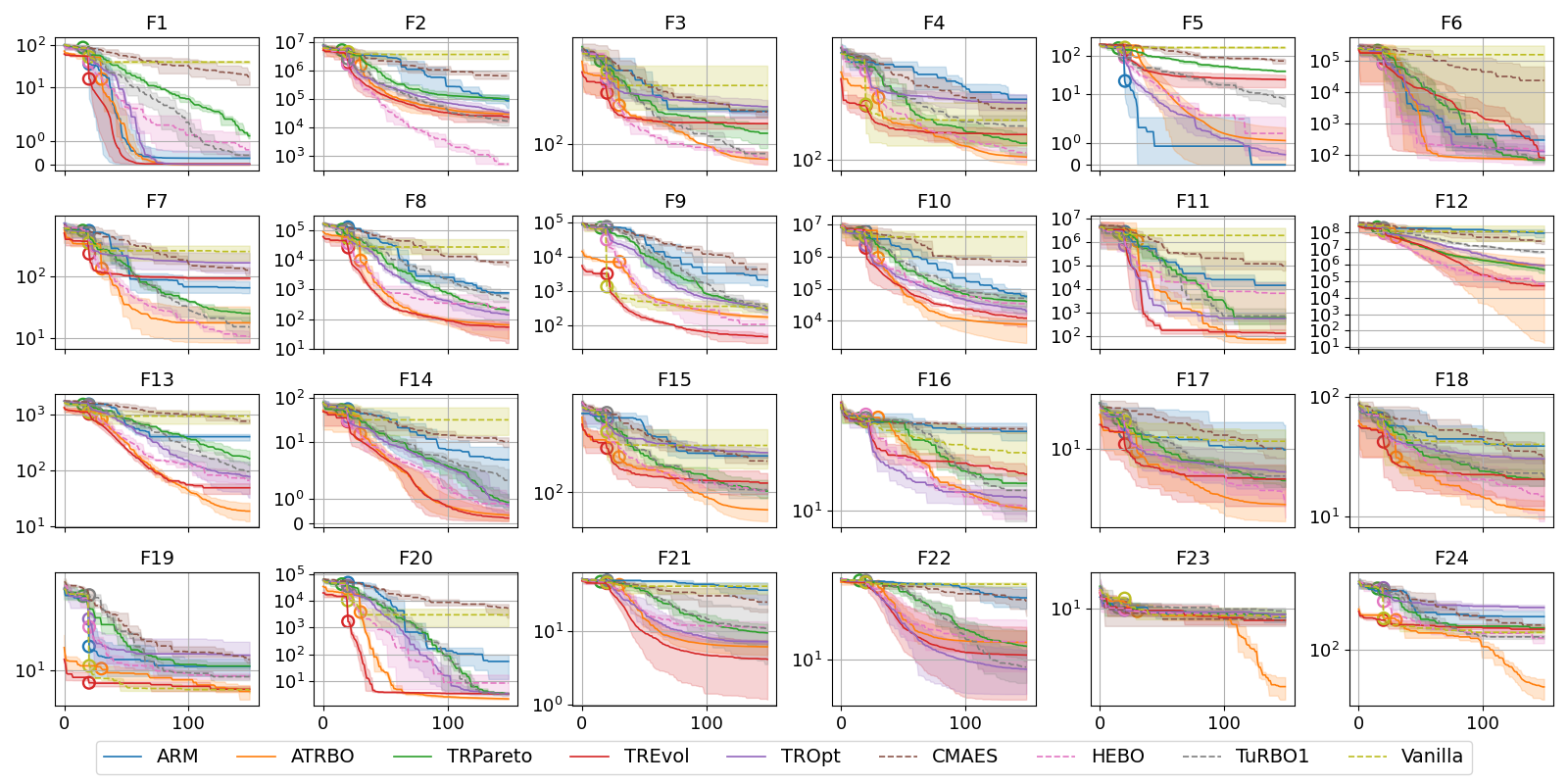}
    \caption{Loss convergence curves on all  BBOB functions in $d=10$.}
    \label{fig:bbob-10d}
\end{figure}

\begin{figure}[H]
    \centering
    \includegraphics[width=\textwidth,trim=0mm 0mm 0mm 0mm,clip]{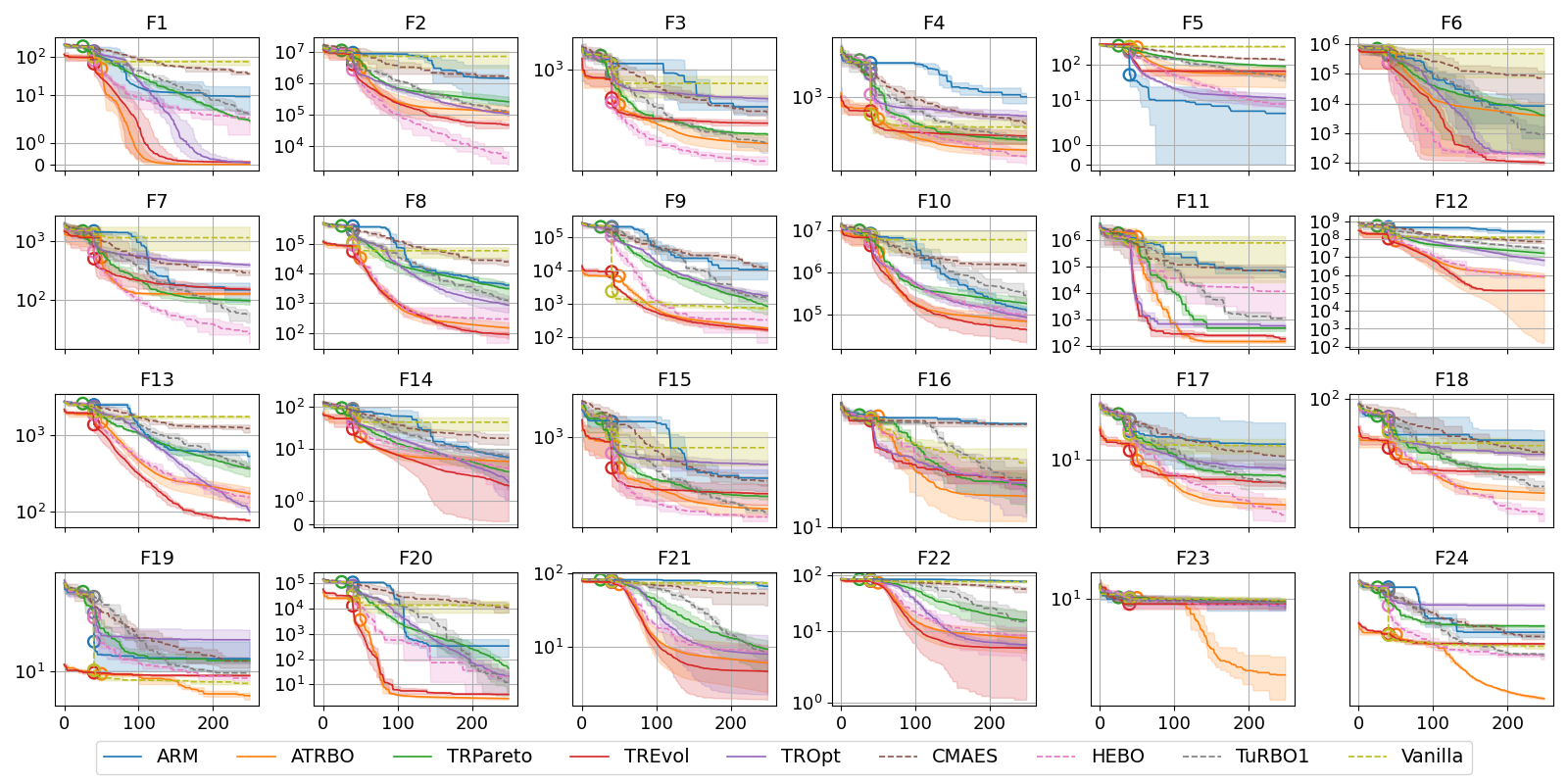}
    \caption{Loss convergence curves on all  BBOB functions in $d=20$.}
    \label{fig:bbob-20d}
\end{figure}

\begin{figure}[H]
    \centering
    \includegraphics[width=\textwidth,trim=0mm 0mm 0mm 0mm,clip]{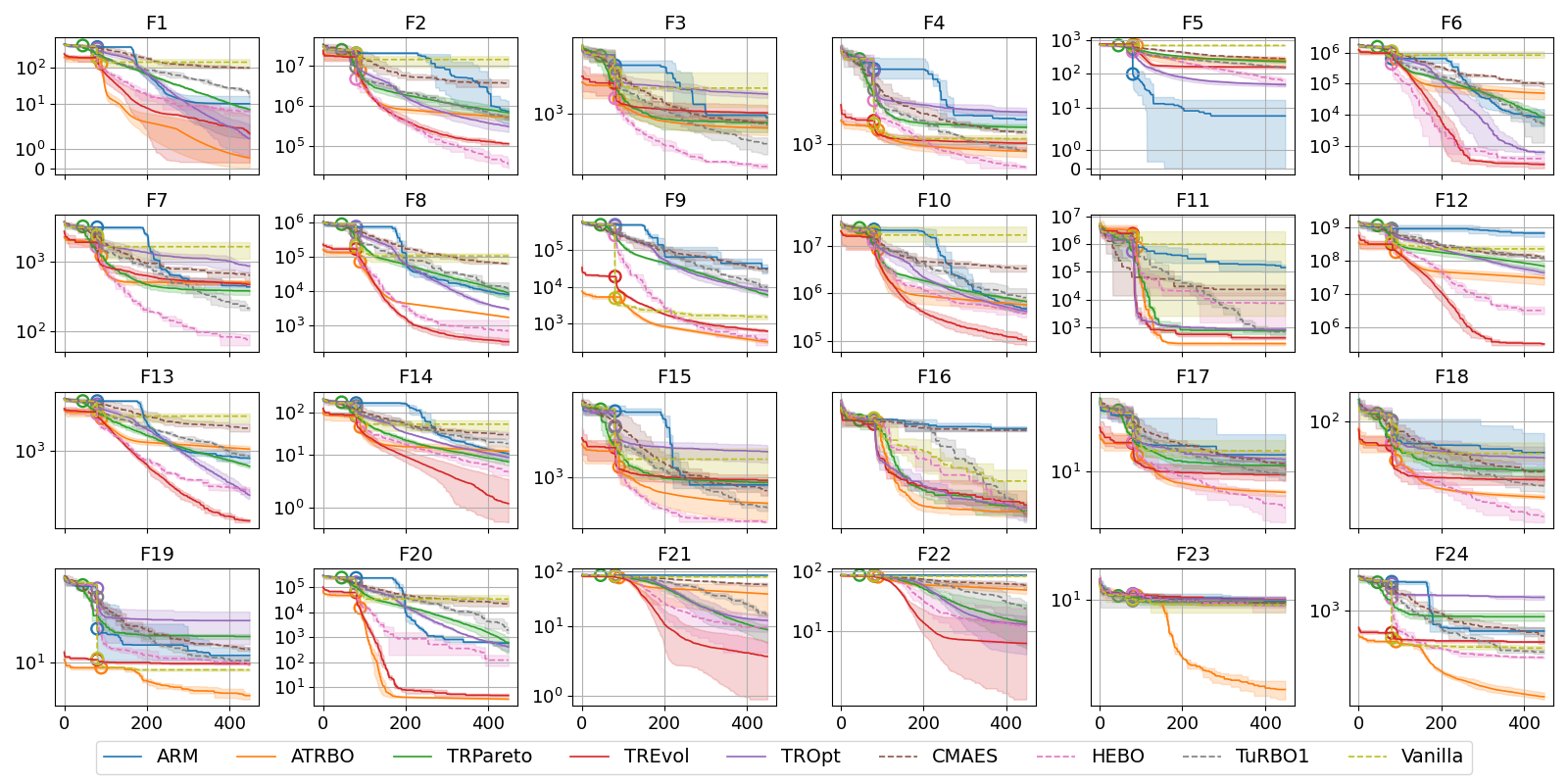}
    \caption{Loss convergence curves on all  BBOB functions in $d=40$.}
    \label{fig:bbob-40d}
\end{figure}

\subsection{Bayesmark}
\label{App:full_results_Bayesmark}

\begin{figure}[H]
    \centering
    \includegraphics[width=\textwidth,trim=0mm 0mm 0mm 0mm,clip]{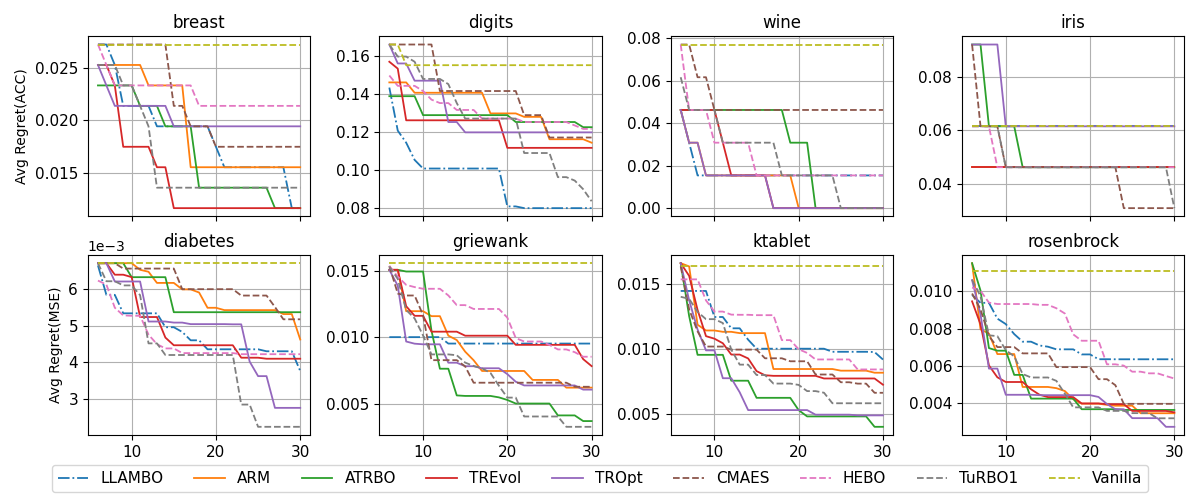}
    \caption{Regret curves on all datasets using Adaboost.}
    \label{fig:bayesmark-ada}
\end{figure}

\begin{figure}[H]
    \centering
    \includegraphics[width=\textwidth,trim=0mm 0mm 0mm 0mm,clip]{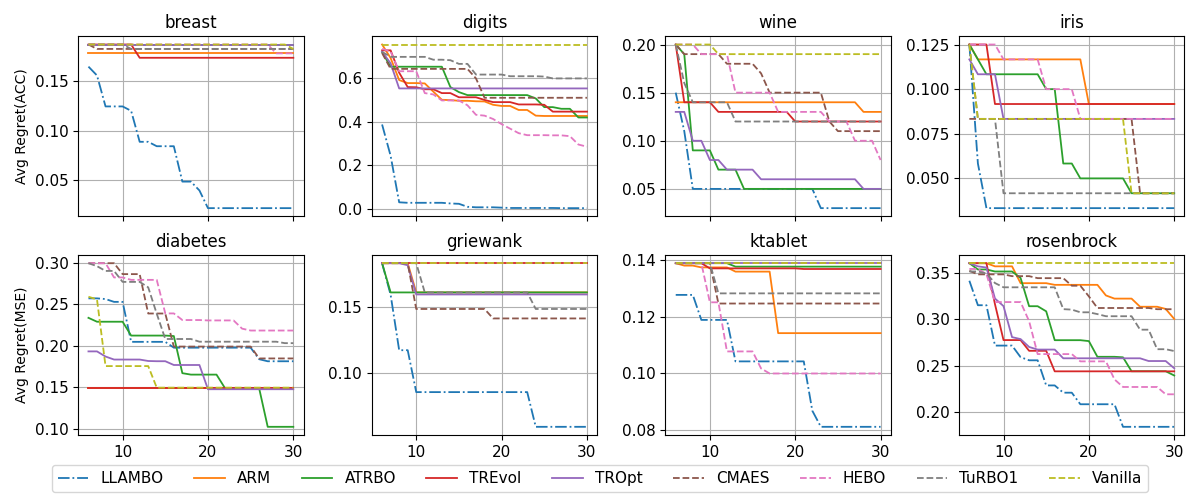}
    \caption{Regret curves on all datasets using Regression Trees.}
    \label{fig:bayesmark-tree}
\end{figure}

\begin{figure}[H]
    \includegraphics[width=\textwidth,trim=0mm 0mm 0mm 0mm,clip]{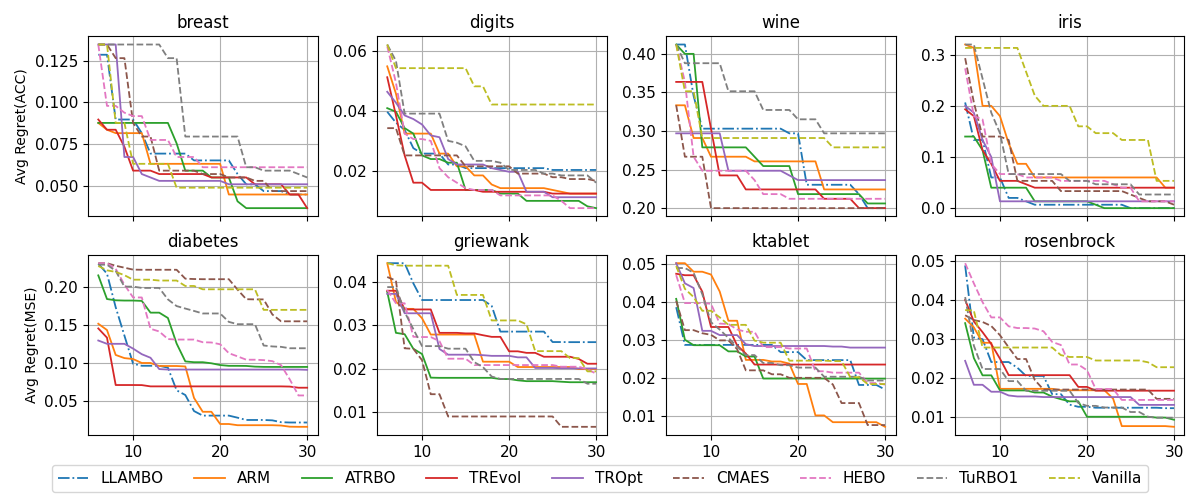}
    \caption{Regret curves on all datasets using MLP SGD.}
    \label{fig:bayesmark-mlp}
\end{figure}

\begin{figure}[H]
    \centering
    \includegraphics[width=\textwidth,trim=0mm 0mm 0mm 0mm,clip]{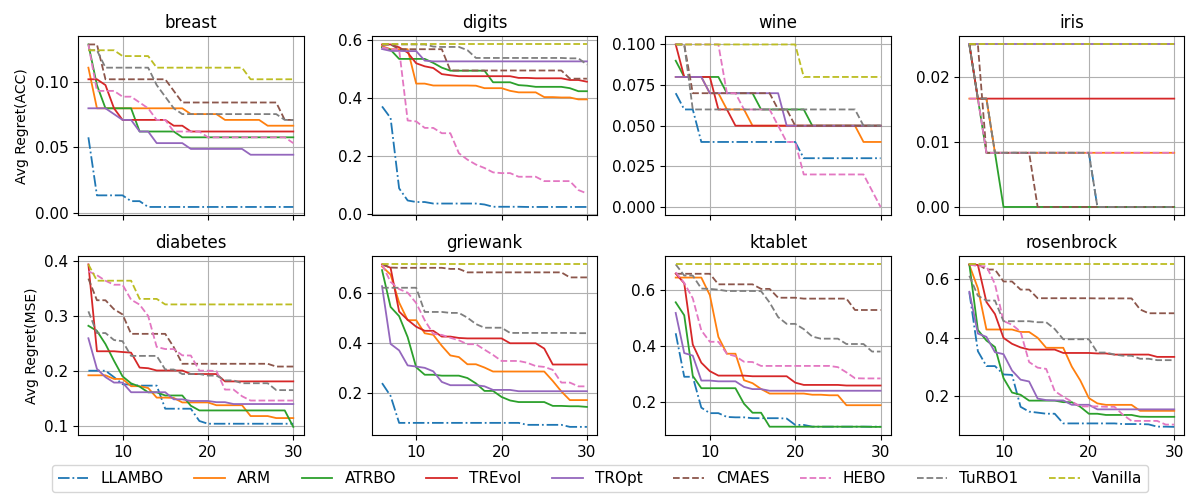}
    \caption{Regret curves on all datasets using Random Forest.}
    \label{fig:bayesmark-rf}
\end{figure}

\begin{figure}[H]
    \centering
    \includegraphics[width=\textwidth,trim=0mm 0mm 0mm 0mm,clip]{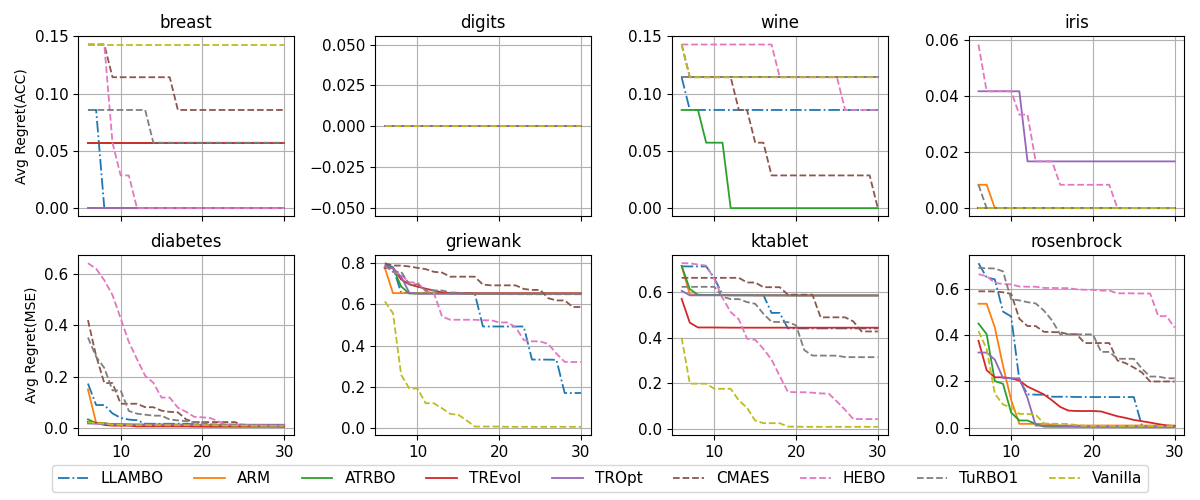}
    \caption{Regret curves on all datasets using SVM.}
    \label{fig:bayesmark-svm}
\end{figure}

\section{CPU Time Analysis}
\begin{figure}[H]
     \centering
     \begin{subfigure}[b]{0.49\textwidth}
         \centering
         \includegraphics[width=\textwidth]{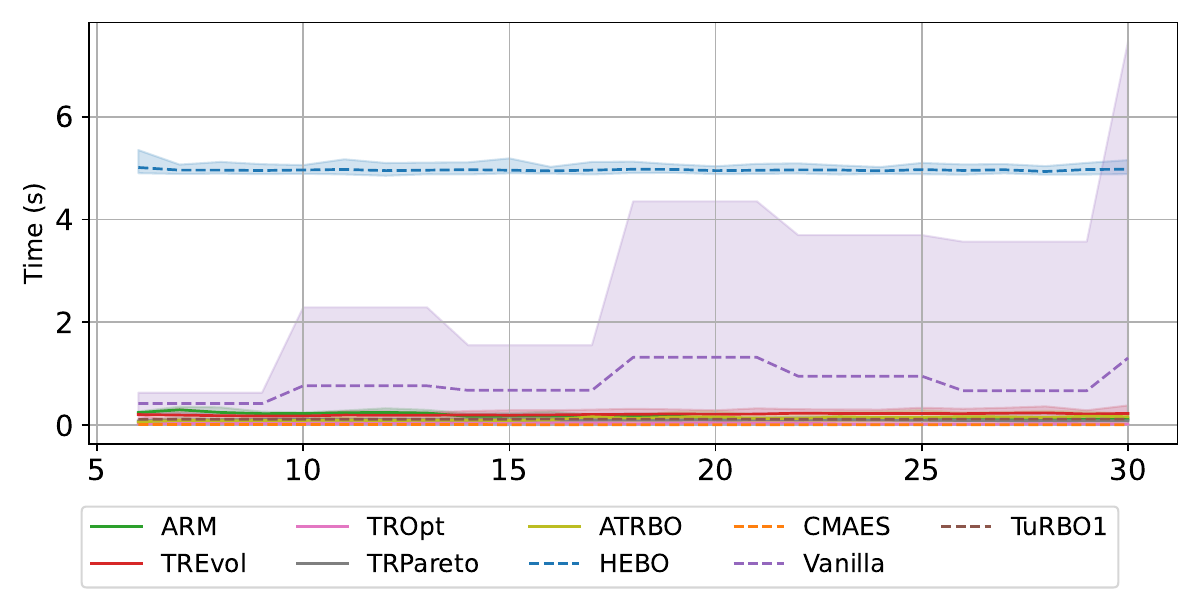}
         \caption{Average CPU time per algorithm.}
     \end{subfigure}
     \hfill
     \begin{subfigure}[b]{0.49\textwidth}
         \centering
        \includegraphics[width=\textwidth]{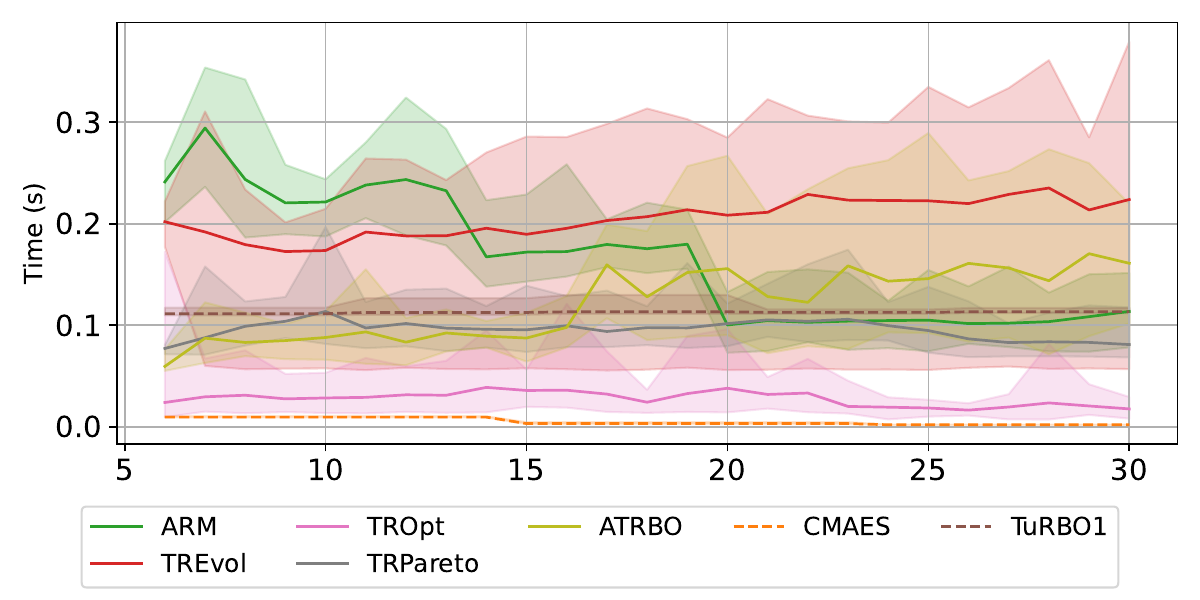}
         \caption{Excluding HEBO and VanillaBO.}
     \end{subfigure}
        \caption{Average CPU time per algorithm on the Random Forest task averaged over 3 datasets (breast, iris, diabetes) and 5 repeats per dataset (15 runs) using an Apple M1 Pro 16 GB consumer laptop.}
        \label{fig:cputime}
\end{figure}

Figure \ref{fig:cputime} shows that generated BO algorithms
run in the same ballpark as \emph{TuRBO1}
($\sim\!0.3$ s on an M1 Pro), and an order of magnitude faster than \textsc{HEBO}.
Note that LLAMBO’s required CPU time as reported by the authors are $3$s (on their hardware) and LLAMBO also requires either a payed LLM API or significant GPU hardware.

\section{Sensitivity Analysis of the Best Algorithm (ATRBO)}

In this section we dissect \emph{ATRBO} to explain its edge and provide good
default values for practitioners. We first provide the pseudo code in Section \ref{app:ATRBO} (full Python code is available in our repository \cite{anonymous_2025_15384611}). We then provide ablation studies for $d=5$ on the most important hyper-parameters of ATRBO in Section \ref{app:hpo}. Lastly we analyze why ATRBO is especially well performing on $f_{23}$ by debugging the different adaptive parameter updates inside the algorithm over the run.

\subsection{Pseudocode of ATRBO}
\label{app:ATRBO}

\input{atrbo_pseudocode}

\subsection{Ablation on the Hyper-parameters}
\label{app:hpo}
\begin{figure}[H]
    \centering
    \includegraphics[width=\textwidth,trim=0mm 0mm 0mm 0mm,clip]{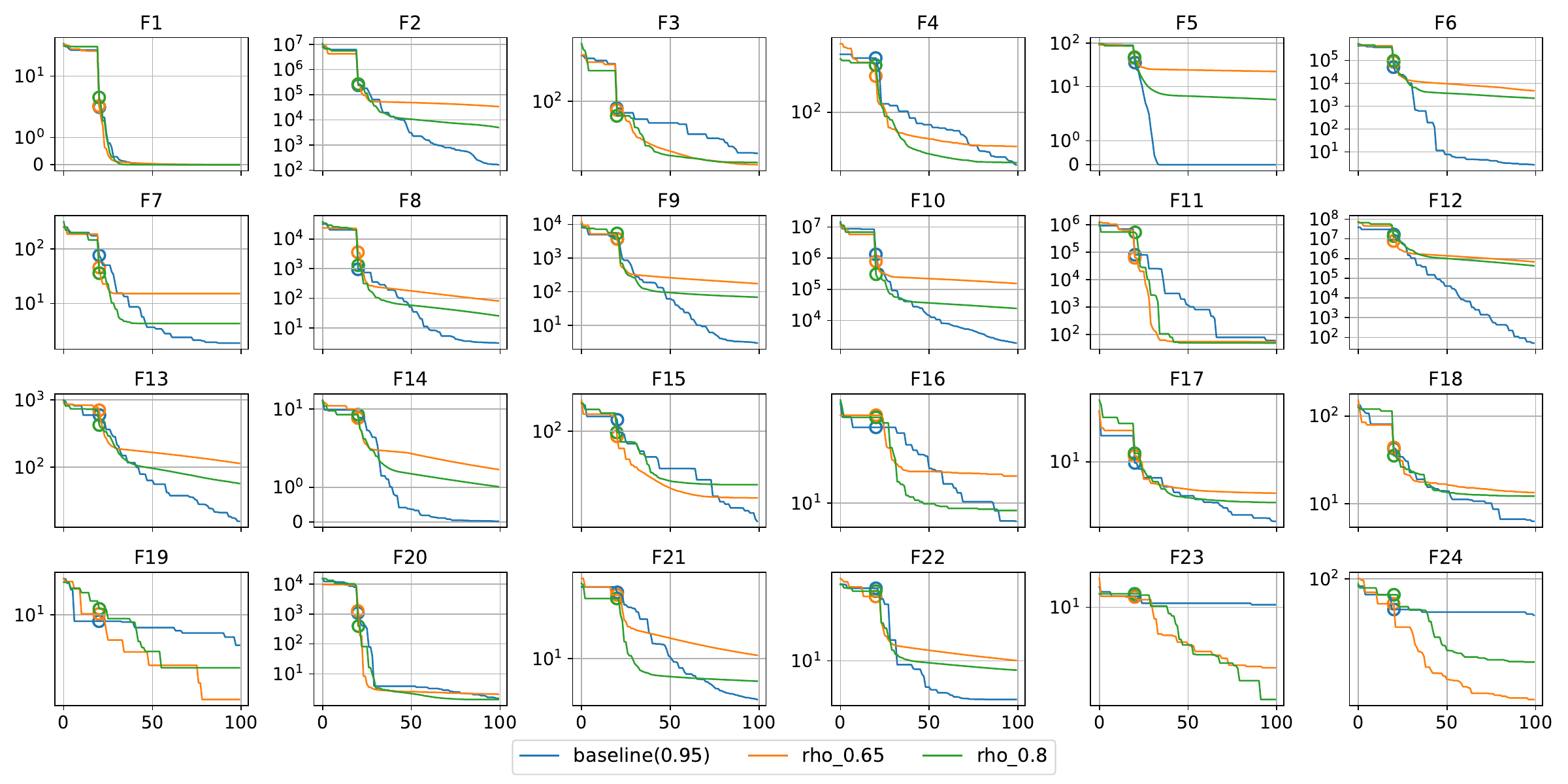}
    \caption{ATRBO $\rho$ settings $[0.65, 0.80, 0.95]$. The baseline has $\rho=0.95$. The convergence curves are averaged over $5$ runs in $d=5$.}
    \label{fig:ablation-atrbo-rho}
\end{figure}

\begin{figure}[H]
    \centering
    \includegraphics[width=\textwidth,trim=0mm 0mm 0mm 0mm,clip]{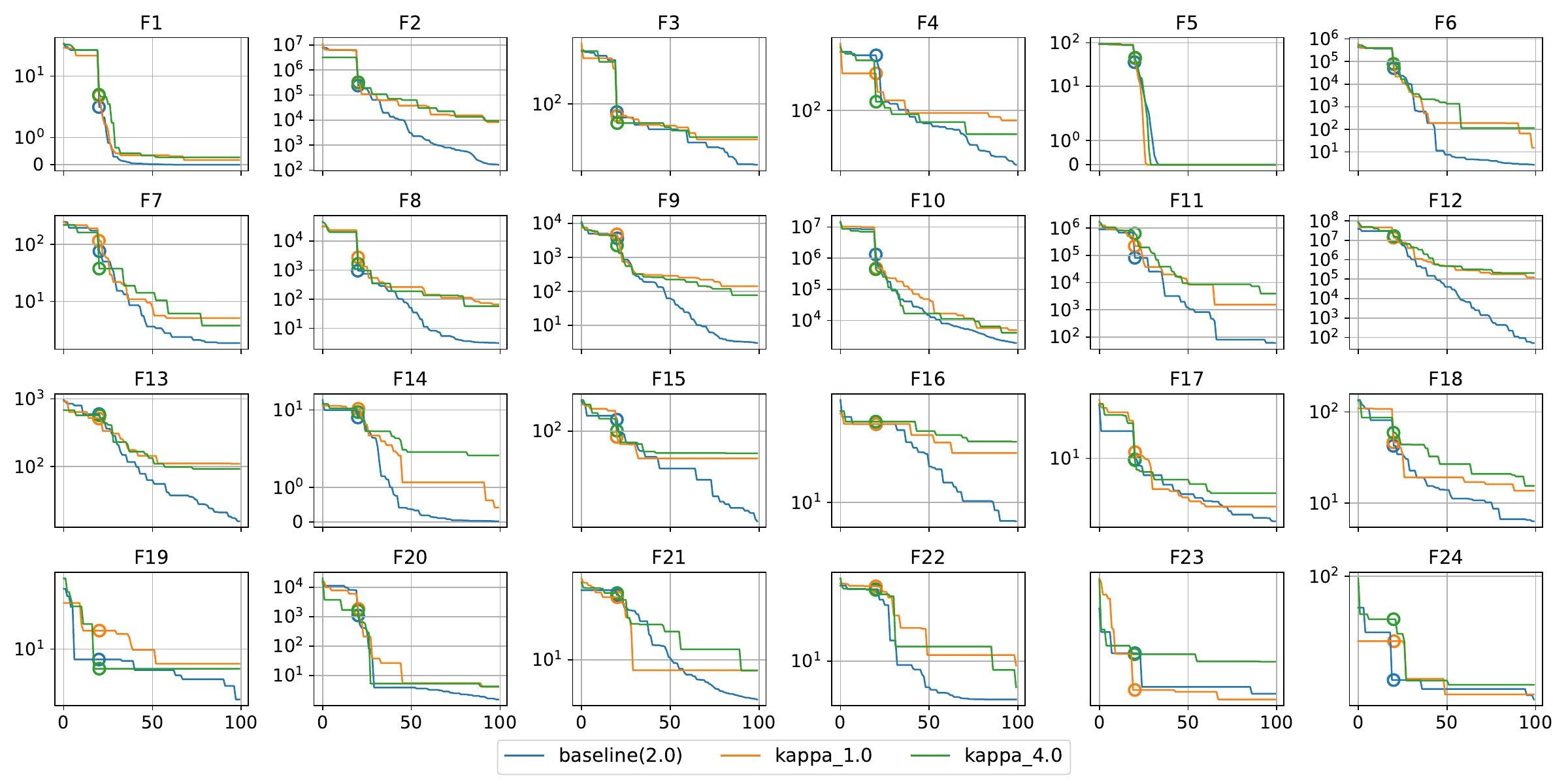}
    \caption{ATRBO $\kappa$ settings $[1.0, 2.0, 4.0]$. The baseline has $\kappa=2.0$. The convergence curves are averaged over $5$ runs in $d=5$.}
    \label{fig:ablation-atrbo-kappa}
\end{figure}

\begin{figure}[H]
    \centering
    \includegraphics[width=\textwidth,trim=0mm 0mm 0mm 0mm,clip]{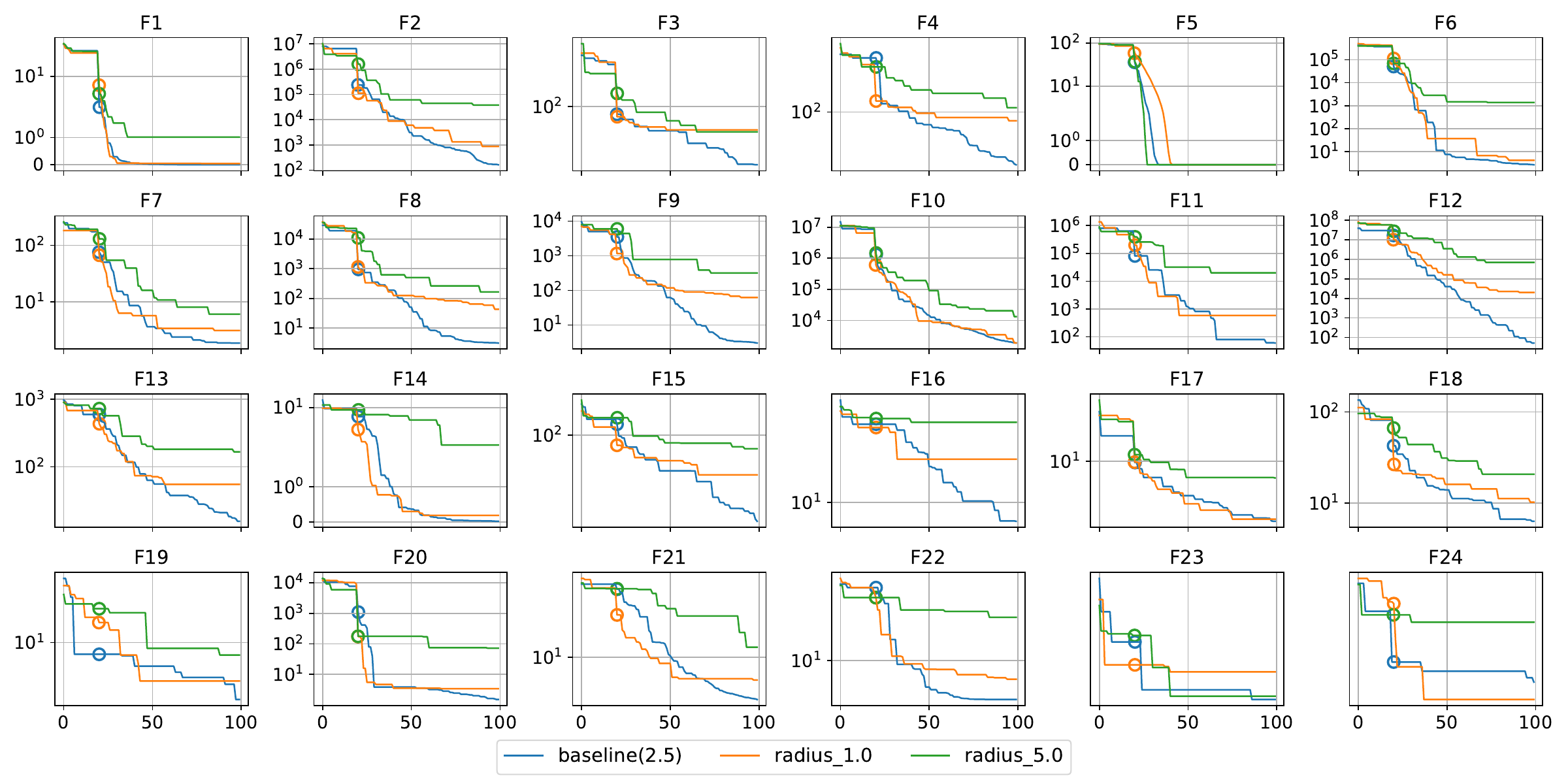}
    \caption{ATRBO radius settings $[1.0, 2.5, 5.0]$. The baseline has radius$=2.5$. The convergence curves are averaged over $5$ runs in $d=5$.}
    \label{fig:ablation-atrbo-radius}
\end{figure}

\begin{figure}[H]
    \centering
    \includegraphics[width=\textwidth,trim=0mm 0mm 0mm 0mm,clip]{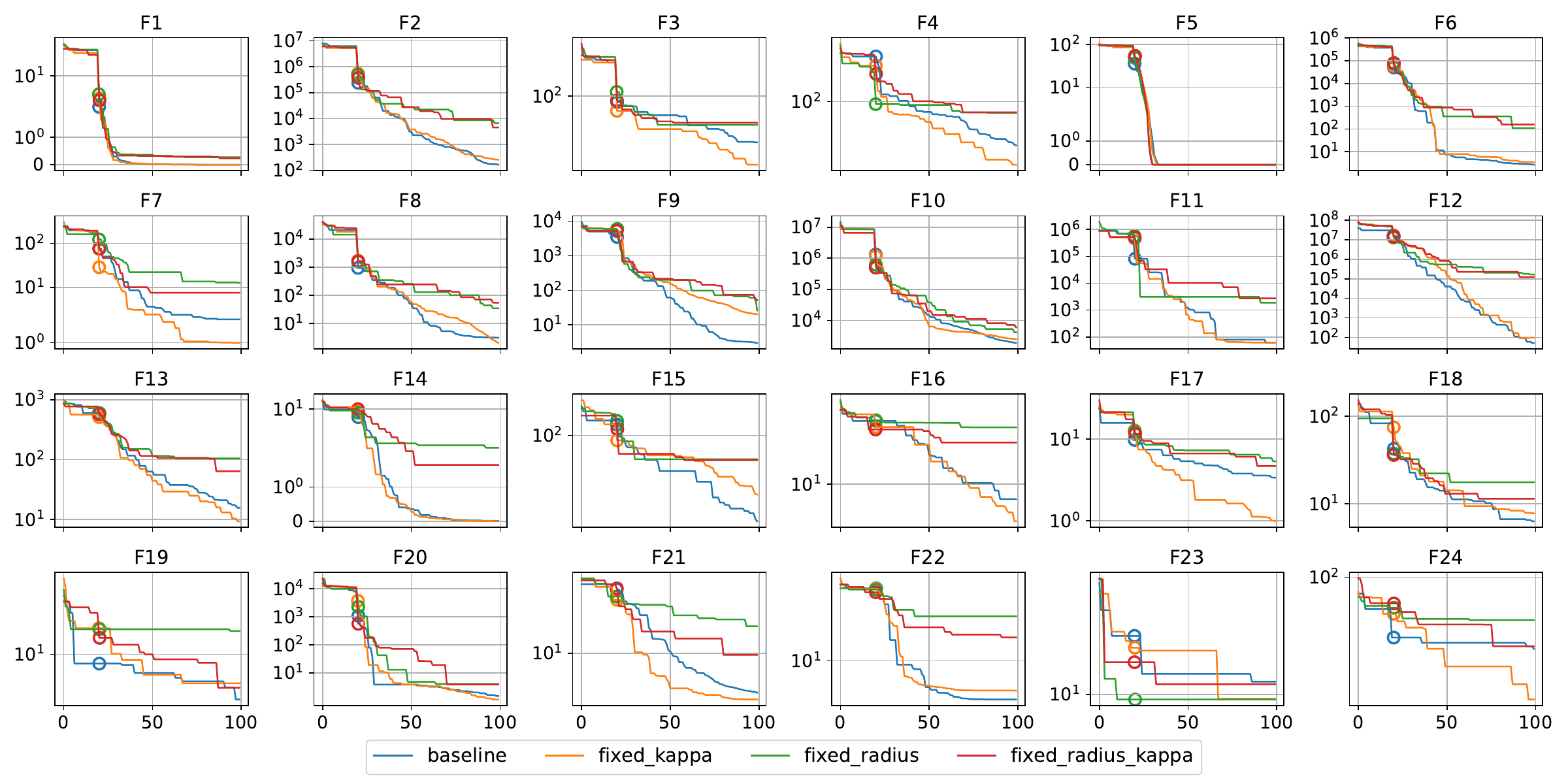}
    \caption{Adaptation modifications, baseline (adaptive $\kappa$ and radius) versus fixed $\kappa$, fixed radius and both fixed. The convergence curves are averaged over $5$ runs in $d=5$.}
    \label{fig:ablation-atrbo-adap}
\end{figure}

The results in the figures below are from ablation studies of the hyper-parameters $\rho$, $\kappa$ and initial radius. The original ATRBO (baseline) is denoted with a blue line. From Figure \ref{fig:ablation-atrbo-rho} we can observe that a value of $0.8$ and $0.65$ can be beneficial for highly multi-modal functions ($f_{23}, f_{24}$), however it degrades the performance in other functions such as $f_{18}$. The results in Figure \ref{fig:ablation-atrbo-kappa} show that the baseline $\kappa$ of $2.0$ is a good choice, different values degrade performance significantly. Similarly, in Figure \ref{fig:ablation-atrbo-radius} shows that the baseline radius of $2.5$ is a god choice overall. Removing the adaptation techniques from the algorithm often degrades the performance (Figure \ref{fig:ablation-atrbo-adap}).

\subsection{Analysis on F23}
\begin{figure}[H]
    \centering
    \includegraphics[width=\textwidth,trim=0mm 0mm 0mm 0mm,clip]{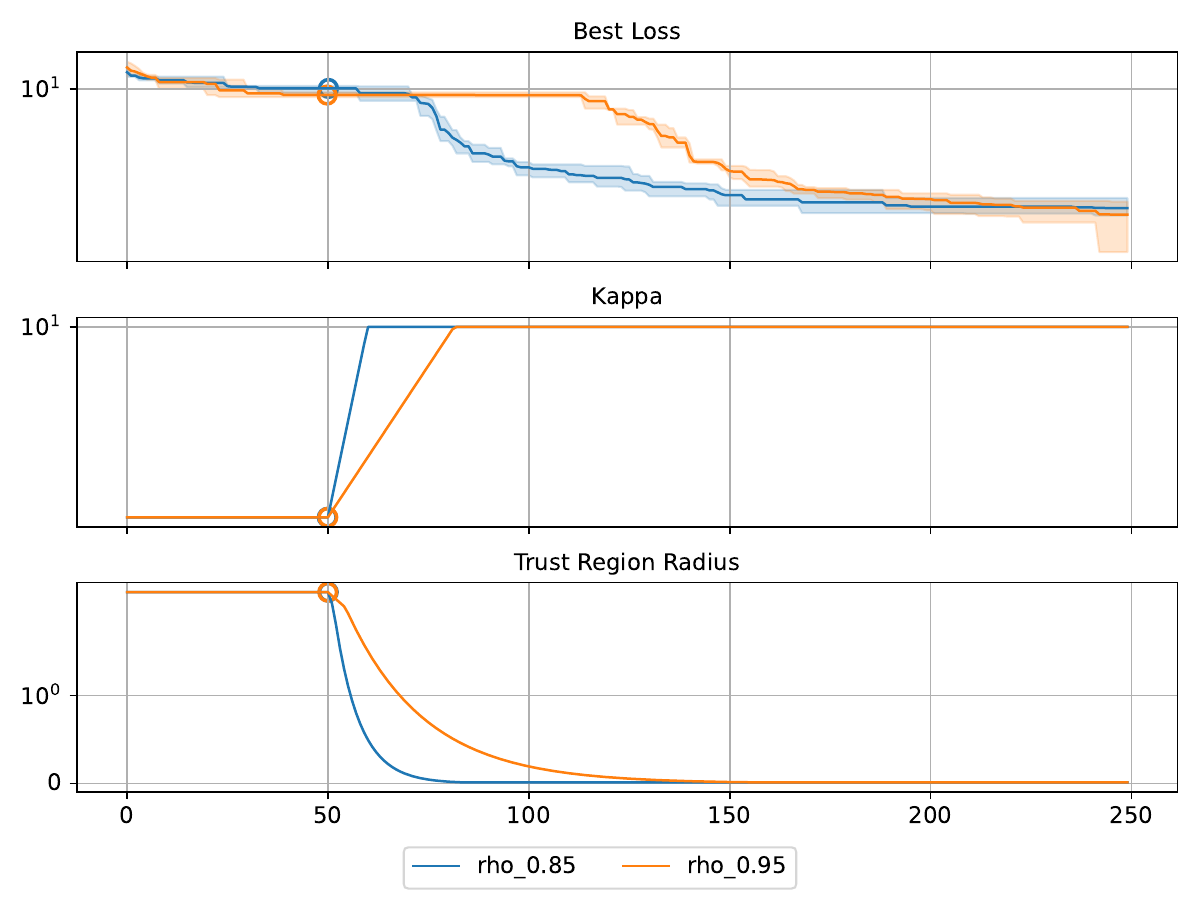}
    \caption{Loss, $\kappa$ and trust region radius over $5$ independent runs with $\rho=0.85$ and $\rho=0.95$ on $f_{23}$ in $20$ dimensions.}
    \label{fig:ablation-atrbo-f23}
\end{figure}

We finally look deeper into the performance on $f_{23}$ as we observed outstanding performance on this very hard to optimize function. On thevery  rugged $f_{23}$ landscape,
Figure \ref{fig:ablation-atrbo-radius} demonstrates that early
radius shrinkage lets ATRBO quickly lock on promising basins, explaining its outstanding performance in high dimensions. From Figure \ref{fig:ablation-atrbo-f23} we can observe that after the initial sampling (design of experiments) of $50$ samples (for $20d$), we see a rapid shrinkage of the trust region radius, which leads to a fast exploitation which is beneficial on $f_{23}$ due to the large number of local minima.

\end{document}

%% file: llameabo_pseudocode.tex
\begin{algorithm}[H]
\algtext*{EndWhile}
\algtext*{EndIf}
\algtext*{EndFor}
\algtext*{EndFunction}

\caption{LLaMEA-BO\label{alg:LLaMEABO}}
\textbf{Input:}\\
\hspace*{\algorithmicindent} $T$: Total number of iterations, $\mu$: Size of parent population\\
\hspace*{\algorithmicindent} $\lambda$: Size of offspring population, $p_{cr}$: Crossover rate\\    
\hspace*{\algorithmicindent} $top\_k$: Top-k parameter for LLM, $temperature$: Temperature parameter for LLM\\
\hspace*{\algorithmicindent} $isElitism$: Boolean indicating whether to use elitism or not.\\
\begin{algorithmic}[1]
\State $t \gets 0$,  $P_t \gets \emptyset$

\Comment{\textit{Initialization Phase}}
\While{$t < \mu$}
    \State $prompt \gets \text{getInitialPrompt}(P_t)$
    \State $solution \gets \text{LLM}(prompt, top\_k, temperature)$
    \State $result \gets \text{Evaluate}(solution)$,  $t \gets t + 1$
    \State $P_t \gets P_t \cup (solution, result)$
\EndWhile

\Comment{\textit{Evolutionary Loop}}
\While{$t < T$}
    \State $prompts \gets \text{getPrompts}(P_t, \min(\lambda, T-t), p_{cr})$
    \State $solutions \gets \text{LLM}(prompts, top\_k, temperature)$
    \State $results \gets \text{Evaluate}(solutions)$
    
    \If{$isElitism$}
        \State $P_t \gets \text{Select}(P_t \cup solutions, results)$ \Comment{Elitist Selection}
    \Else
        \State $P_t \gets \text{Select}(solutions, results)$ \Comment{Non-Elitist Selection}
    \EndIf

    \State $t \gets t + |P_t|$
\EndWhile

\tikzmk{A} 

\Function{getPrompts}{$P$, $size$, $p_{cr}$}
    \State $sortedP \gets \text{sort}(P)$ \Comment{Sort population by fitness}
    \State $Pa_{cr} \gets \text{comb}(sortedP)$ \Comment{Create combinations for crossover}
    \State $Pa_{\mu} \gets sortedP$
    \State $prompts \gets \emptyset$
    \For{$i \gets 0$ \textbf{to} $size-1$}
        \If{$\text{rand}() < p_{cr}$}
            \State $parents \gets \text{dequeue}(Pa_{cr})$ \Comment{Perform crossover}
        \Else
            \State $parents \gets \text{dequeue}(Pa_{\mu})$ \Comment{Perform mutation}
        \EndIf
        \State $prompts \gets prompts \cup \text{generatePrompt}(P, parents)$

        \If{$|Pa_{cr}| = 0$}
            \State $Pa_{cr} \gets \text{comb}(sortedP)$
        \EndIf
        \If{$|Pa_{\mu}| = 0$}
          \State $Pa_{\mu} \gets sortedP$
        \EndIf
    \EndFor
    \State \Return $prompts$
\EndFunction

\tikzmk{B} 
\boxit{mycyan}

\end{algorithmic}
\end{algorithm}

%% file: atrbo_pseudocode.tex
\begin{algorithm}[H]
\algtext*{EndWhile}
\algtext*{EndIf}
\algtext*{EndFor}
\algtext*{EndFunction}
\caption{ATRBO \label{alg:ATRBO}}
\begin{algorithmic}[1]
\Require Budget $B$, Dimension $d$
\State Initialize:
    \State \hspace{0.5cm} $X \gets \emptyset$, $y \gets \emptyset$ \Comment{Observed points and values}
    \State \hspace{0.5cm} $n_{init} \gets \min(10d, B/5)$ \Comment{Number of initial exploration points}
    \State \hspace{0.5cm} $x_{best} \gets \text{None}$, $y_{best} \gets \infty$ \Comment{Best point and value found}
    \State \hspace{0.5cm} $r \gets 2.5$ \Comment{Trust region radius}
    \State \hspace{0.5cm} $\rho \gets 0.95$ \Comment{Radius adjustment factor}
    \State \hspace{0.5cm} $\kappa \gets 2.0$ \Comment{Exploration-exploitation parameter}
    \State \hspace{0.5cm} $n_{evals} \gets 0$
    \State \hspace{0.5cm} Bounds $\gets$ $[[-5, ..., -5], [5, ..., 5]]$

\Function{SamplePoints}{$n$, $c$, $r$} \Comment{Sample $n$ points around center $c$ within radius $r$}
    \State \hspace{0.5cm} Sample $n$ points from Sobol sequence scaled to $[-1, 1]^d$
    \State \hspace{0.5cm} Project points to hypersphere of radius 1
    \State \hspace{0.5cm} Scale points to radius $r$ and center at $c$
    \State \hspace{0.5cm} Clip points to within Bounds
    \State \hspace{0.5cm} \Return Scaled points
\EndFunction

\Comment{Initialization}
\State $X_{init} \gets$ \Call{SamplePoints}{$n_{init}$, Bounds.mean(), Bounds.range()}
\State $y_{init} \gets$ Evaluate($f$, $X_{init}$), $n_{evals} \gets n_{evals} + X_{init}$
\State $x_{best} \gets X[\arg \min y]$

\Comment{Optimization Loop}
\While{$n_{evals} < B$}
    \State $GP \gets$ Fit($X$, $y$)
    \State $X_{samples} \gets$ \Call{SamplePoints}{$100d$, $x_{best}$, $r$} \Comment{Sample points for acquisition function}
    \State $acq \gets$ LCB($X_{samples}$, $GP$, $\kappa$) \Comment{Lower Confidence Bound}
    \State $x_{next} \gets X_{samples}[\arg \min acq]$
    \State $y_{next} \gets$ Evaluate($f$, $x_{next}$), $n_{evals} \gets n_{evals} + 1$
    \State $x_{best} \gets X[\arg \min y]$
    \State $y_{best} \gets \min y$

    \Comment{Adjust Trust Region}
    \State $r \gets r * \rho$
    \State $\kappa \gets \kappa / \rho$

    \State $r \gets \text{clip}(r, 10^{-2}, \max(Bounds.range())/2)$
    \State $\kappa \gets \text{clip}(\kappa, 0.1, 10.0)$

\EndWhile

\State \textbf{return} $y_{best}$, $x_{best}$
\end{algorithmic}
\end{algorithm}